%% file: main.tex
\documentclass{IEEEtaes}

\usepackage{amsmath,amsfonts,amssymb}
\usepackage{color, array, amsthm}
\usepackage{graphicx}
\usepackage{booktabs}
\usepackage{xcolor}
\usepackage{url}

\usepackage{cite}
\usepackage{hyperref} 
\usepackage{url} 
\usepackage{subcaption} 
\usepackage{cleveref} 
\usepackage{todonotes} 
\usepackage{placeins} 
\usepackage{ragged2e}

\jvol{XX}
\jnum{XX}
\jmonth{XXXXX}
\paper{1234567}
\pubyear{2022}
\doiinfo{TAES.2022.Doi Number}

\begin{document}

\title{CageDroneRF: A Large-Scale RF Benchmark and Toolkit for Drone Perception}

\author{Mohammad Rostami\textsuperscript{*}%
}
\affil{Department of Electrical and Computer Engineering at Rowan University, Glassboro, NJ, USA} 

\author{Atik Faysal\textsuperscript{*}}
\member{Member, IEEE}
\affil{Department of Electrical and Computer Engineering at Rowan University, Glassboro, NJ, USA} 

\author{Hongtao Xia}
\affil{AeroDefense, Oceanport, NJ, USA}

\author{Hadi Kasasbeh}
\affil{AeroDefense, Oceanport, NJ, USA}

\author{Ziang Gao}
\affil{AeroDefense, Oceanport, NJ, USA}

\author{Huaxia Wang}
\member{Member, IEEE}
\affil{Department of Electrical and Computer Engineering at Rowan University, Glassboro, NJ, USA}

\receiveddate{Manuscript received XXXXX 00, 0000; revised XXXXX 00, 0000; accepted XXXXX 00, 0000.}

\corresp{Corresponding author: H. Wang \textsuperscript{*}Equal contribution.}

\authoraddress{Mohammad Rostami, Atik Faysal, and Huaxia Wang are with the Department of Electrical and Computer Engineering at Rowan University, Glassboro, NJ, USA (e-mail: \href{mailto:rostami23@rowan.edu}{rostami23@rowan.edu}; \href{mailto:faysal24@rowan.edu}{faysal24@rowan.edu}; \href{mailto:wanghu@rowan.edu}{wanghu@rowan.edu}).\\
Hongtao Xia, Hadi Kasasbeh, and Ziang Gao are with AeroDefense, Oceanport, NJ, USA (e-mail: \href{mailto:hongtao.xia@aerodefense.tech}{hongtao.xia@aerodefense.tech}; \href{mailto:hadi.kasasbeh@aerodefense.tech}{hadi.kasasbeh@aerodefense.tech}; \href{mailto:ziang.gao@aerodefense.tech}{ziang.gao@aerodefense.tech}).}

\editor{The CDRF dataset, code, and trained models are publicly available at \href{https://github.com/DroneGoHome/U-RAPTOR-PUB}{https://github.com/DroneGoHome/U-RAPTOR-PUB}.\\
\href{https://aerodefense.tech/u-raptor-data-request/}{https://aerodefense.tech/u-raptor-data-request}}

\supplementary{Color versions of one or more of the figures in this article are available online at \href{http://ieeexplore.ieee.org}{http://ieeexplore.ieee.org}.}

\markboth{ROSTAMI ET AL.}{CDRF: A LARGE-SCALE RF BENCHMARK AND TOOLKIT FOR DRONE DETECTION}

\maketitle

\input{sections/0_abstract}

\input{sections/1_introduction}

\input{sections/2_related_works}
\input{sections/3_dataset_design}

\input{sections/4_experiments_results}

\input{sections/5_conclusion}

\FloatBarrier
\begingroup
\justifying
\bibliographystyle{IEEEtaes.bst} 
\bibliography{bibtex/bib/IEEEexample} 
\endgroup
\end{document}

%% file: sections/0_abstract.tex
\begin{abstract}
    {We present CageDroneRF (CDRF), a large-scale benchmark for Radio-Frequency (RF) drone detection and identification built from real-world captures and systematically generated synthetic variants. CDRF addresses the scarcity and limited diversity of existing RF datasets by coupling extensive raw recordings with a principled augmentation pipeline that (i)~precisely controls Signal-to-Noise Ratio (SNR), (ii)~injects interfering emitters, and (iii)~applies frequency shifts with label-consistent bounding-box recomputation for detection. The dataset spans a wide range of contemporary drone models, many of which are unavailable in current public datasets, and diverse acquisition conditions, derived from data collected at the Rowan University campus and within a controlled RF-cage facility. CDRF is released with interoperable open-source tools for data generation, preprocessing, augmentation, and evaluation that also operate on existing public benchmarks. It enables standardized benchmarking for classification, open-set recognition, and object detection, supporting rigorous comparisons and reproducible pipelines. By releasing this comprehensive benchmark and tooling, we aim to accelerate progress toward robust, generalizable RF perception models.}
\end{abstract}


%% file: sections/1_introduction.tex
\section{Introduction}
\label{sec:introduction}
{The rapid proliferation of Unmanned Aerial Vehicles (UAVs) has enabled transformative applications in logistics, agriculture, inspection, emergency response, and transportation \cite{robotics12020053}, while simultaneously introducing acute risks to safety, privacy, and critical infrastructure \cite{nguyen2025effective}. High-profile disruptions near airports, increasing reports of unauthorized flights over sensitive facilities, and the use of UAVs in illicit operations underscore the operational need for reliable, scalable counter-UAV systems \cite{basak2021combined}. Early and effective detection and identification are essential components of any response strategy~\cite{allahham2020deep}.}

{Traditional sensing modalities such as radar, vision, and acoustics exhibit complementary strengths but also notable failure modes. Radar can struggle with small targets at long range, vision systems depend on favorable lighting and line-of-sight conditions, and acoustic sensors suffer from limited detection range and high sensitivity to ambient noise. All three can incur substantial deployment costs. Radio-Frequency (RF) sensing offers distinct advantages. Because most consumer and professional drones maintain continuous RF links for command-and-control and video transmission, RF-based systems can detect and characterize targets day or night, in non-line-of-sight scenarios, and at comparatively low cost. Moreover, RF captures are intrinsically rich for learning, carrying device, protocol, and operator signatures across diverse channels~\cite{mo2022deep}.}

{Despite these advantages, progress in Machine Learning (ML) for RF-based drone perception has lagged behind other domains. A central barrier is the absence of broad, standardized benchmarks and tools. Existing datasets are often narrow in class diversity and limited in raw volume; many are collected in relatively clean environments with narrow Signal-to-Noise Ratio (SNR) ranges and weak interference diversity. As a result, models trained on such data can overfit to idealized conditions, achieving near-perfect scores on easy benchmarks yet degrading severely in realistic deployments with Bluetooth/Wi-Fi interference, spectrum crowding, and low SNR. The cost and logistics of collecting sufficiently varied RF captures further slow iteration and reproducibility~\cite{rfuav}.}

We introduce CageDroneRF (CDRF), a benchmark built to close this gap with a dataset-and-toolkit co-design. CDRF comprises real-world captures from controlled RF-cage facilities and open-campus settings, paired with a principled raw‑signal augmentation pipeline that programmatically varies SNR, injects interfering emitters, and applies frequency shifts while preserving label consistency. Crucially, augmentations operate on complex baseband I/Q, before time-frequency conversion, so synthesized conditions faithfully reflect RF phenomena rather than image-level artifacts. For detection, frequency shifts are accompanied by exact recalculation of You-Only-Look-Once (YOLO~\cite{redmon2016you})-format annotations with correct wrap-around behavior on the spectrogram’s frequency axis. The data processing stack converts raw captures to spectrograms via Short-Time Fourier Transform (STFT) and exposes all parameters (e.g., sampling rate, FFT size, segment length) for reproducible ablations and reprocessing.

{These design choices yield several key differentiators over existing RF drone benchmarks, discussed in detail in Section~\ref{subsec:comparison}. Briefly, CDRF spans 39~classes across 23~drone models with dual-environment collection (Faraday cage and outdoor), adopts a 20\,MHz sampling rate chosen for edge-device feasibility, and, most critically, maintains a complete, parameterized processing chain from raw I/Q through augmentation to detection annotation: any signal-level transformation automatically yields recomputed spectrogram images and bounding-box labels, an end-to-end capability absent from all prior RF drone datasets.}

{Beyond the dataset, CDRF provides interoperable utilities for dataset creation, cleaning, and evaluation. The data module loads \texttt{.dat} files via memory mapping, slices long recordings into time windows, generates spectrograms, appends per-sample metadata (including time bounds), and optionally adds controlled AWGN, Rician, or Rayleigh fading noise for precise SNR targets. The YOLO toolkit includes raw-IQ augmentation and automatic bounding-box recomputation; a dataset cleaner aligns third-party releases (e.g., Roboflow-style~\cite{roboflow_website}) into a consistent hierarchy; and a lightweight patch exposes per-detection class probabilities from YOLO for richer calibration and analysis. For classification, CDRF ships PyTorch~\cite{paszke2019pytorch} baselines (binary and multi-class) built on ResNet-18~\cite{he2016deep} with ready-to-run dataloaders for both spectrogram images and array/pickle formats, standardizing training, metrics, and confusion matrices. All tooling is dataset-agnostic, enabling its application to existing public datasets and facilitating cross-benchmark comparability.}

{Collectively, CDRF aims to shift the field from narrow, clean benchmarks toward realistic, stress-tested evaluation under varied SNRs, interference, and frequency offsets, all of which are conditions that operational systems routinely encounter. By releasing both data and the underlying signal-first augmentation/evaluation stack, we seek to catalyze reproducible research on RF-based detection, identification, and related perception tasks. Our key contributions are as follows:}
\begin{itemize}
    \item \textbf{CDRF benchmark dataset:} Real-world RF captures from cage environments with standardized spectrogram generation and rich per-sample metadata.
    {\item \textbf{Raw-signal augmentation pipeline:} Controlled SNR injection, interfering-signal mixing, and frequency shifts applied at I/Q level, with exact YOLO label recomputation (including wrap-around).
    \item \textbf{SNR-structured datasets:} Programmatic creation of SNR-stratified splits (including noise-only backgrounds) for stress testing and realistic robustness evaluation.
    \item \textbf{Interoperable tooling:} Open-source utilities for dataset creation, metadata generation, Roboflow-style cleaning, and spectrogram rendering; tools are dataset-agnostic for use on existing releases.}
    \item \textbf{Detection analysis enhancements:} Patch exposing per-detection class probabilities from YOLO to support calibration studies and open-set analyses.
    \item \textbf{Baselines and evaluation:} Ready-to-run PyTorch baselines for binary and multi-class classification with standardized metrics and confusion matrices, enabling reproducible benchmarking across tasks.
    \item \textbf{Open set recognition:} Implementation of open-set recognition capabilities to handle previously unseen classes during inference.
\end{itemize}

The remainder of this paper is organized as follows. Section~\ref{sec:related_work} reviews related work in RF-based drone detection, covering various sensing modalities and existing public datasets. Section~\ref{sec:dataset} details the design and collection of the CDRF dataset, and Section~\ref{sec:data_processing} describes our data preprocessing, augmentation, and annotation pipeline. Section~\ref{sec:ml_tasks} presents the experimental setup and benchmark results for several ML tasks, including drone detection, single-label, open-set, and hierarchical classification. Finally, Section~\ref{sec:conclusion} concludes the paper and discusses future research directions.

%% file: sections/2_related_works.tex
\section{Related Work}
\label{sec:related_work}

The landscape of drone detection and classification has evolved rapidly under the pressure of monitoring and mitigating unauthorized UAVs. Research spans multiple sensing modalities, including radar, acoustics, vision, and RF, each with distinct strengths and failure modes. Below, we summarize these modalities, review RF signal representations and learning methods, and discuss public datasets that have shaped the field, highlighting persistent gaps that motivate the development of a new benchmark and tooling.

\subsection{Drone Detection Methodologies}
\label{subsec:modalities}
\begin{itemize}
    \item \textbf{Radar-based} \cite{basak2021combined,fioranelli2015classification, 
    drozdowicz201635, coluccia2020detection, zhu2020classification, roldan2020dopplernet, rahman2020classification, moses2011radar, mendis2016deep, park2015combination
    } systems exploit micro-Doppler and related motion signatures and can operate at long range and in adverse weather, but often struggle with small, slow, or low Radar Cross-Section (RCS) targets and occlusions.
    \item \textbf{Acoustic-based} \cite{andravsi2017night,mezei2015drone,mezei2016drone, nijim2016drone,yue2018software,bernardini2017drone, park2015combination, liu2017drone,kim2017real,casabianca2021acoustic
    } methods rely on propeller/engine signatures and are low-cost and lightweight, yet are highly sensitive to ambient noise and typically limited to short ranges.
    \item \textbf{Vision-based} \cite{liu2017drone,gokcce2015vision,saqib2017study,thomas2019uav} systems (RGB/thermal/multispectral) can excel at fine-grained recognition in favorable conditions but degrade with poor lighting, weather, or occlusions, and often require high-resolution optics and line-of-sight.
    \item \textbf{RF-based} \cite{rfuav,al2019rf,al2020drone,akter2020rf,medaiyese2021machine,taha2019machine,nguyen2018cost} sensing leverages command-and-control and video links that persist during operation, enabling day/night, non-line-of-sight detection with modest hardware costs and rich device/protocol signatures.
    {\item \textbf{Multi-Modality} \cite{akter2022explainable, jamil2020malicious, svanstrom2021real, diamantidou2019multimodal,liu2017drone, jovanoska2018multisensor,mahjourian2024multimodal}, often referred to as multi-sensor fusion, involves the integration of data from two or more distinct sensing modalities to enhance drone detection and classification capabilities. This approach is widely regarded as a crucial strategy for building truly robust and comprehensive counter-drone systems.}
\end{itemize}

To contextualize the role and trade-offs of RF sensing within the broader landscape of drone detection, Table~\ref{tab:sensor_comparison} provides a comparative analysis of major sensor modalities.

\begin{table*}[!ht]
\centering
\caption{Comparison of Sensor Modalities for Drone Detection}
\label{tab:sensor_comparison}
\renewcommand{\arraystretch}{1.3} 
\begin{tabular}{|p{0.15\textwidth}|p{0.3\textwidth}|p{0.3\textwidth}|c|}
\hline
\textbf{Detection Technology} & \textbf{Primary Advantage (Pros)} & \textbf{Primary Disadvantage (Cons)} & \textbf{Citations} \\
\hline
\textbf{Radio Frequency (RF)} & Highly advantageous and cost-effective. Unaffected by weather or lighting conditions. Can operate in Non-Line-of-Sight (NLoS) scenarios. Can detect specific command and control frequencies and offer early warnings. & Extremely sensitive to co-channel interference (e.g., Wi-Fi, Bluetooth). Not perfect for autonomous drones (which may lack continuous RF communication). Limited performance at low SNR. & \cite{nguyen2025effective,seidaliyeva2023advances,teoh2019rf,mrabet2024machine,al2024deep}\\
\hline
\textbf{Visual/Optical} & Provides high-accuracy identification and is relatively low cost. Often utilizes effective ML techniques like Convolutional Neural Networks (CNNs). & Performance is severely degraded by adverse weather and poor lighting. Requires a clear Line-of-Sight (LoS). High false positive rates due to resemblance to birds or other objects. & \cite{mrabet2024machine, medaiyese2022wavelet,basak2021combined,taha2019machine}\\
\hline
\textbf{Radar} & Offers long detection ranges. Less affected by weather and time of day. Provides precise location data. & Conventional radar struggles with mini-drones due to their low Radar Cross-Section (RCS). High expense and regulatory licensing required. Can cause RF interference. & \cite{basak2021combined,mrabet2024machine,al2024deep} \\
\hline
\textbf{Acoustic} & Cost-effective and passive operation. Can be miniaturized. & Extreme susceptibility to ambient noise. Very short detection range. Cannot detect quieter drones utilizing noise reduction. & \cite{medaiyese2022wavelet, bello2019radio,al2024deep}\\
\hline
\textbf{LiDAR} & Offers higher spatial resolution than radar for precise 3D mapping. Excels in low-light or adverse weather. & Requires direct LoS. More expensive and power-intensive than radar. Relatively short range. & \cite{mrabet2024machine}\\
\hline
\textbf{Thermal/Infrared} & Can detect heat emitted by objects, performing well in night surveillance and adverse weather. & Prohibitively expensive for long distances. Performs poorly on drones with plastic bodies or electric motors that radiate less heat. May mistake hotter objects for drones. & \cite{kilicc2022drone, bello2019radio}\\
\hline
\textbf{Multi-Sensor Fusion} & Overcomes the inherent limitations of any single sensor by leveraging complementary strengths. Provides enhanced accuracy and reliability, reducing false alarms. & Increased complexity of system design. Requires advanced sensor fusion techniques to combine data from different formats. & \cite{mrabet2024machine,akter2022explainable,diamantidou2019multimodal}\\
\hline
\end{tabular}
\end{table*}

\subsection{RF Signal Transformation and Feature Representations}
\label{subsec:rf_representation}
A central challenge in RF-based perception is learning discriminative features from raw I/Q streams. Visual time-frequency representations have become a standard connection to modern ML.
\begin{itemize}
    \item \textbf{Spectrograms via STFT} provide localized energy distributions over time and frequency, exposing cues such as center frequency, bandwidth, dwell time, and hop rate that correlate with device and mode. They have proven effective inputs for Convolutional Neural Networks (CNNs) and detectors.
    \item \textbf{Power Spectral Density (PSD)} offers frequency content but discards temporal dynamics; \textbf{scalograms} (wavelets~\cite{grossmann1984decomposition}) and \textbf{Wigner–Ville distributions}~\cite{cohen1995time} trade off resolution, cross-terms, and interpretability.
    \item \textbf{Alternative encodings}, such as Frequency-Domain Gramian Angular Fields (FDGAF)~\cite{bai2020intelligent}, map 1D spectra into 2D images to better preserve amplitude/temporal structure; \textbf{wavelet scattering}~\cite{bruna2013invariant} has also shown promise for bias removal and transient capture on RF benchmarks.
\end{itemize}

{Empirical evidence suggests that spectrogram-based models are more robust than raw 1D I/Q pipelines under low SNR and co-channel interference, particularly in Industrial, Scientific, and Medical (ISM) bands congested by Wi-Fi and Bluetooth~\cite{rfuav, fu2024radio, medaiyese2022wavelet}.}

\subsection{Deep Learning for RF-based Drone Detection}
\label{subsec:rf_dl}
Deep learning models dominate modern RF perception pipelines:
\begin{itemize}
    \item \textbf{Deep Neural Networks (DNNs)} \cite{abeywickrama2018rf, shijith2017breach,kim2016drone,allahham2020deep} achieved early success for binary detection but often degrade with many classes or confusable spectra.
    \item \textbf{CNNs} \cite{KILIC2022101028, basak2021drone} on spectrograms or raw I/Q deliver strong accuracy; purpose-built architectures (e.g., CNN-SSDI \cite{akter2021cnn}, FDGAF-CNN\cite{fu2024radio}) report high performance on DroneRF-style tasks.
    \item \textbf{Object detectors} \cite{rfuav,dadrass2022modified, kabir2021deep} (e.g., YOLO) jointly localize and classify RF emissions on spectrograms, offering center-frequency and bandwidth estimates alongside class labels; lightweight variants target real-time operation.
    \item \textbf{Residual networks} \cite{podder2024deep} (ResNet-18/50~\cite{he2016deep}) serve as strong spectrogram classifiers; more recent {Transformers~\cite{vaswani2017attention}} (e.g., Swin~\cite{liu2021swin}, ViT~\cite{dosovitskiy2020image}), {EfficientNet~\cite{tan2019efficientnet}}, and {MobileNet~\cite{howard2017mobilenets}} have also been explored for accuracy–efficiency trade-offs.
    \item \textbf{Sequence models} \cite{frid2024drones} (e.g., Long Short-Term Memory (LSTM)~\cite{hochreiter1997long}, Temporal CNNs~\cite{lea2016temporal}) can leverage burst timing; multimodal fusion (RF+acoustics) has shown benefits at low SNR.
    \item \textbf{Classical ML} \cite{drone2024daiarticle} (e.g., XGBoost~\cite{chen2016xgboost} on engineered features) remains competitive in binary detection on clean datasets but tends to be less robust under severe interference.
\end{itemize}

A recurring theme is the sensitivity of performance to data realism: models trained on clean, low-interference datasets show inflated benchmark scores yet generalize poorly under spectrum crowding and low SNR \cite{elyousseph2024robustness,gluge2024robust}.

\subsection{Public Datasets for RF-based Drone Detection}
\label{subsec:datasets}
Public datasets have catalyzed progress but reveal a consistent realism gap:
\begin{itemize}
    \item {\textbf{DroneRF} \cite{allahham2019dronerf} provided early stimulus for deep learning with three drone models, multiple operational modes, and magnitude-only spectra collected in a controlled environment.}
    \item {\textbf{DroneDetect / DroneDetect V2} \cite{5jjj-1m32-21} expanded device diversity to seven drones and explicitly included co-channel interference (Wi-Fi, Bluetooth, combined), providing raw complex I/Q captures recorded via a Nuand BladeRF SDR with GNURadio.}
    \item {\textbf{Cardinal RF} \cite{1xp7-ge95-22} targets classification amid significant interference across drones and non-drone emitters.}
    \item {\textbf{VTI\_DroneSET\_FFT} \cite{VTI_DroneSET_FFT} covers three DJI models across operational modes with intentional Wi-Fi/Bluetooth clutter, distributed in preprocessed \texttt{.mat} format.}
    \item {\textbf{Noisy Drone RF Signal Dataset} \cite{gluge2023robust} offers standardized benchmarking across synthetically controlled SNR ranges (e.g., $-20$ to $30$\,dB), distributed as preprocessed tensors.}
    \item {\textbf{RFUAV} \cite{rfuav} is the largest existing benchmark, comprising $\sim$1.3\,TB of raw complex I/Q from 37~UAV types collected in real-world settings with USRPs, accompanied by XML metadata and open preprocessing tooling with utilities for synthesizing controlled SNR conditions.}
\end{itemize}

{Despite this growing landscape, the utility of existing datasets is constrained by persistent challenges: (i)~limited diversity in device classes and environmental conditions; (ii)~restricted access to raw I/Q signals, preventing custom feature extraction; (iii)~poorly characterized or narrow SNR distributions; (iv)~a scarcity of large-scale, real-world negative samples; and (v)~the absence of an integrated framework for augmentation, robust annotation handling, and benchmarked evaluation.}

\subsection{{Comparison with Prior Datasets and Motivation for CDRF}}
\label{subsec:comparison}

{To concretely motivate the design of CDRF, we provide a unified comparison with the benchmarks reviewed in Section~\ref{subsec:datasets}, organized along five axes: class and environment diversity, data format and raw I/Q accessibility, hardware feasibility, raw-to-annotation traceability, and tooling.}

{\textbf{Class and environment diversity.} Existing datasets cover a narrow range of drone models and collection conditions. DroneRF~\cite{allahham2019dronerf} includes only three drone models in a clean controlled environment, producing near-ceiling benchmark accuracy that transfers poorly to operational settings. DroneDetect~V2~\cite{5jjj-1m32-21} expands to seven models with real-world interference but provides insufficient background-only segments for training well-calibrated binary detectors. VTI\_DroneSET\_FFT~\cite{VTI_DroneSET_FFT} covers three DJI models. Cardinal~RF~\cite{1xp7-ge95-22} suffers from restricted public availability, limiting reproducibility. RFUAV~\cite{rfuav} is the largest with 37~UAV types but its captures are predominantly high-SNR, with low-SNR and interference conditions synthesized post-hoc rather than captured in situ. In contrast, CDRF spans 39~classes across 23~drone models and combines Faraday-cage isolation for clean reference captures with open-campus outdoor recordings under natural interference, a dual-environment methodology absent from all prior datasets.}

{\textbf{Data format and raw I/Q accessibility.} DroneRF provides only magnitude-based spectral profiles, precluding any baseband-level processing. VTI\_DroneSET\_FFT distributes preprocessed \texttt{.mat} files. Noisy Drone RF Signal Dataset~\cite{gluge2023robust} distributes preprocessed tensors. In all three cases, researchers cannot apply novel signal-processing or augmentation pipelines because raw I/Q is unavailable. DroneDetect~V2 and RFUAV do provide raw I/Q, but as discussed below, providing raw recordings alone is insufficient without a pipeline that preserves the link to annotations.}

{\textbf{Hardware feasibility.} CDRF adopts a 20\,MHz sampling rate deliberately chosen for edge-device compatibility. By contrast, RFUAV's 100\,MHz (100\,MS/s) complex sampling rate, while maximizing spectral coverage, imposes computational and memory demands that are impractical for real-time, resource-constrained deployments where signal capture, spectrogram generation, and inference must all execute within tight budgets.}

{\textbf{Raw-to-annotation traceability.} This distinction is, in our view, the most consequential. In existing benchmarks that distribute raw I/Q recordings alongside pre-generated spectrogram images and annotations (e.g., RFUAV, DroneDetect~V2), there is no automated, reproducible link between the raw signal and the image-level labels. If a researcher modifies the I/Q recording, for instance by injecting noise, applying a frequency shift, or mixing an interferer, the pre-existing annotations become invalid with no means to recover them: the spectrogram must be regenerated from the modified signal, and every bounding box must be re-annotated manually. The same problem arises whenever the spectrogram generation parameters themselves are changed: adjusting the sampling rate (e.g., to target a different hardware platform), FFT size, window length, or hop size alters both the time and frequency resolution of the resulting image, so every pixel coordinate, and therefore every bounding box, changes. Because prior datasets provide only fixed, pre-rendered spectrograms and their corresponding labels, any such parameter change forces a complete manual re-annotation of the entire dataset. This effectively renders the raw I/Q data read-only for any task that requires detection-level labels. CDRF eliminates this barrier by maintaining a complete, parameterized processing chain from raw I/Q through spectrogram generation to detection annotation. When a signal-level transform is applied or the spectrogram parameters are modified, the pipeline automatically produces a new spectrogram image \emph{and} recomputes the corresponding bounding-box annotations, including correct wrap-around handling for frequency shifts. This end-to-end traceability enables unlimited programmatic generation of new, correctly labeled training samples from any raw recording under any chosen set of processing parameters, a capability that no prior RF drone dataset provides.}

{\textbf{Tooling and evaluation infrastructure.} None of the above benchmarks ships dataset-agnostic tooling for cross-benchmark evaluation, nor do they expose per-detection class probability vectors for calibration and open-set analyses. CDRF provides interoperable open-source utilities for dataset creation, metadata generation, cleaning, and evaluation that also operate on existing public datasets, standardizing reproducible benchmarking across the field.}

{Collectively, the field is shifting from model-centric to data- and system-centric evaluation, demanding benchmarks that scale in class and environment diversity, explicitly structure SNR and interference conditions, release raw I/Q for reproducible reprocessing, and ship tooling that enforces label-consistent transforms and standardized evaluation. CDRF is designed to meet each of these requirements.}

%% file: sections/3_dataset_design.tex
\section{CDRF: Design and Collection}
\label{sec:dataset}

{The CDRF dataset is designed to cover the RF signals emitted by the most common UAV-related devices on the market, including downlink video signals and uplink Remote Controller (RC) signals. We begin by recording UAV signals in a relatively clean RF environment to emphasize the characteristics of the signals of interest and reduce the impact of interfering RF sources and environmental variations. Clean data also simplifies data preprocessing and subsequent augmentation.}

\subsection{Data Acquisition Platform}
\begin{figure}[!t]
    \centering
    \begin{subfigure}[h]{0.45\columnwidth}
        \centering
        \includegraphics[width=\columnwidth]{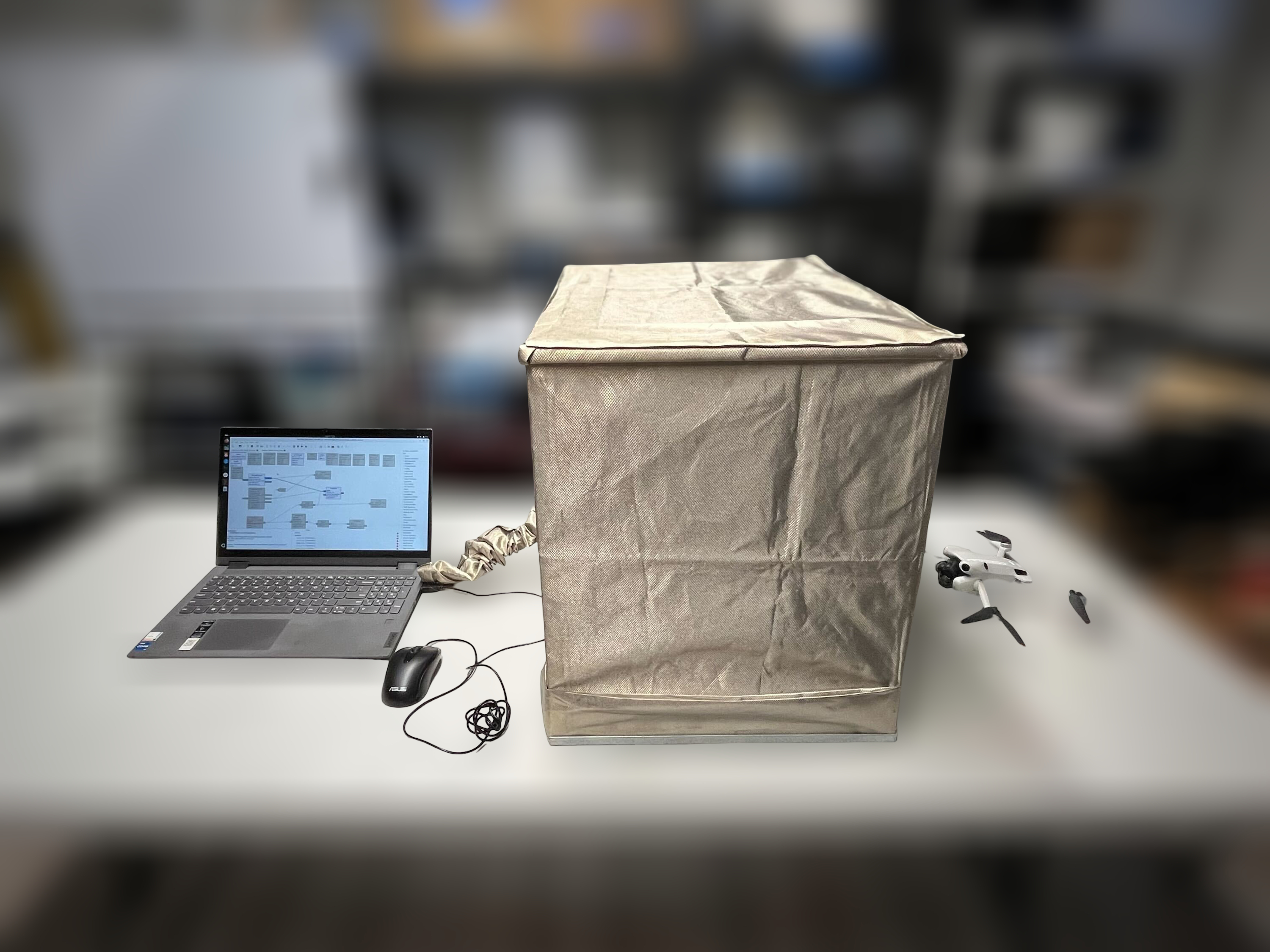}
        \caption{Recording configuration.}
    \end{subfigure}
    \hfill
    \begin{subfigure}[h]{0.45\columnwidth}
        \centering
        \includegraphics[width=\columnwidth]{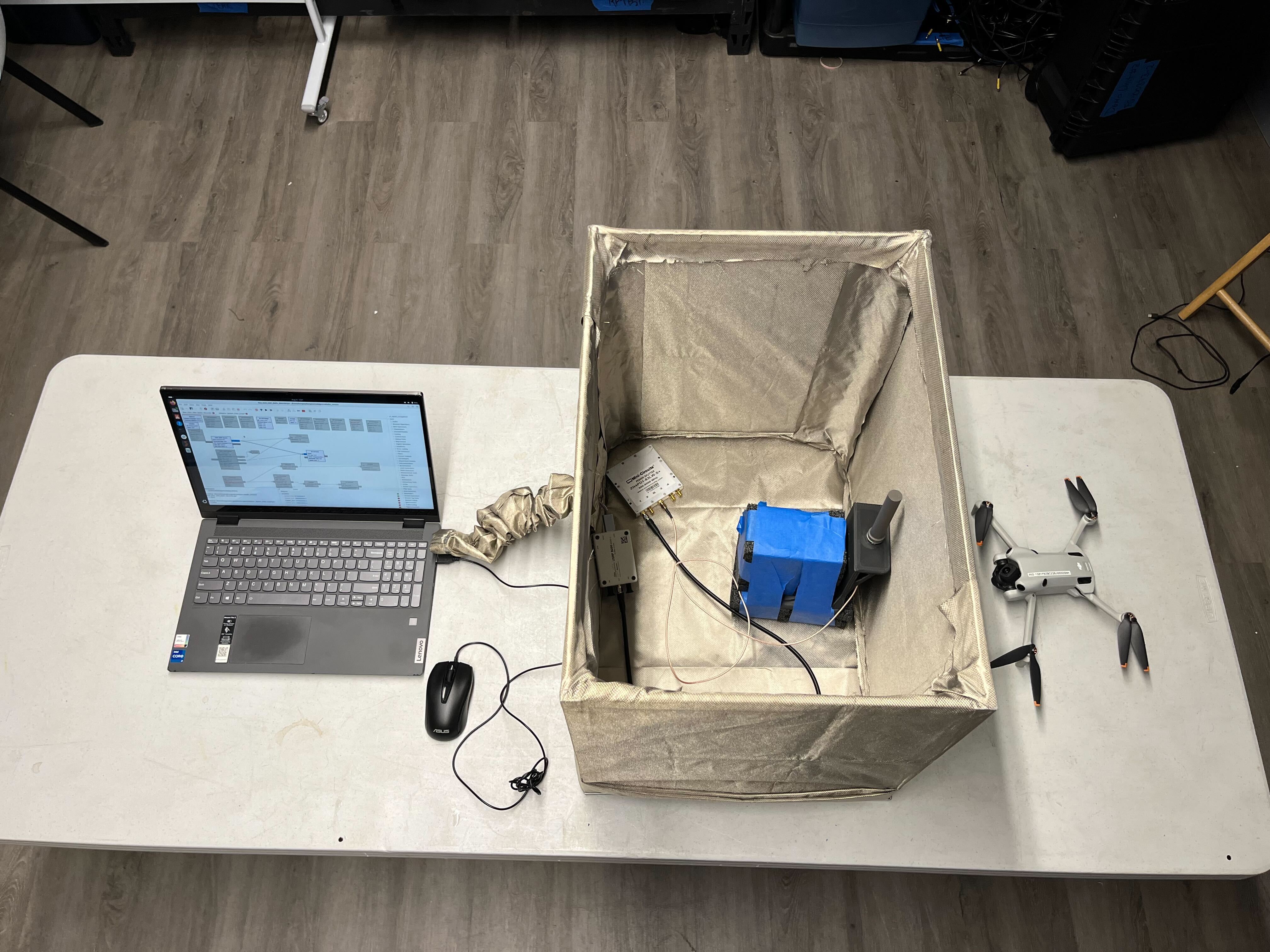}
        \caption{Recording equipment.}
    \end{subfigure}
    \caption{Equipment used to capture the data: portable RF shielded enclosure, SDR card, a laptop, and a drone.}
    \label{fig:equipments}
\end{figure}

{We built the data collection platform using a Software-Defined Radio (SDR) based system. To create an RF-isolation platform, we constructed a customized Faraday cage and placed both the SDR card and receiving antenna inside the cage to attenuate interfering RF signals. The hardware setup and platform configuration are shown in Fig.~\ref{fig:equipments}.}

We use a Lenovo IdeaPad Flex 5 as the data sink. The machine is equipped with an 8-core Intel Core i7-1165G7 CPU and 16 GB of memory. The operating system running on the machine is Ubuntu 24.04.2 LTS. The data collection flow and scripts are built through GNU Radio v3.10 to bridge the Universal Software Radio Peripheral (USRP) B200-mini for RF signal receiving and processing. The SDR device is then connected to a dual-band omnidirectional antenna, which is attached to a customized mount unit to ensure the antenna is static during data collection. 

To ensure comprehensive coverage of drone transmissions, we employ sweep scanning across relevant frequency bands during data collection. This approach allows us to detect and record signals from devices operating at variable frequencies.
    
\subsection{Data Collection Methodology}
{To initiate the data collection, we set up the platform as described in Fig.~\ref{fig:equipments}(a), then power on the UAV and its binding controller. While leaving the antenna inside the cage, we place the UAV outside at a very close distance to the antenna. This placement ensures that the UAV signals delivered to the antenna are strong, while other interfering RF sources in the surrounding area are at greater distances and their signals are largely blocked by the Faraday cage. In addition, the RC of the UAV is kept in an adjacent room. Our goal is to minimize the RC signals captured by the collection platform in this phase so that the dataset can support separate analysis of UAV and RC signals. This separation also provides greater flexibility for subsequent data augmentation.} 

\begin{figure}[!t]
    \centering
    \begin{subfigure}[h]{0.45\columnwidth}
        \includegraphics[width=\columnwidth]{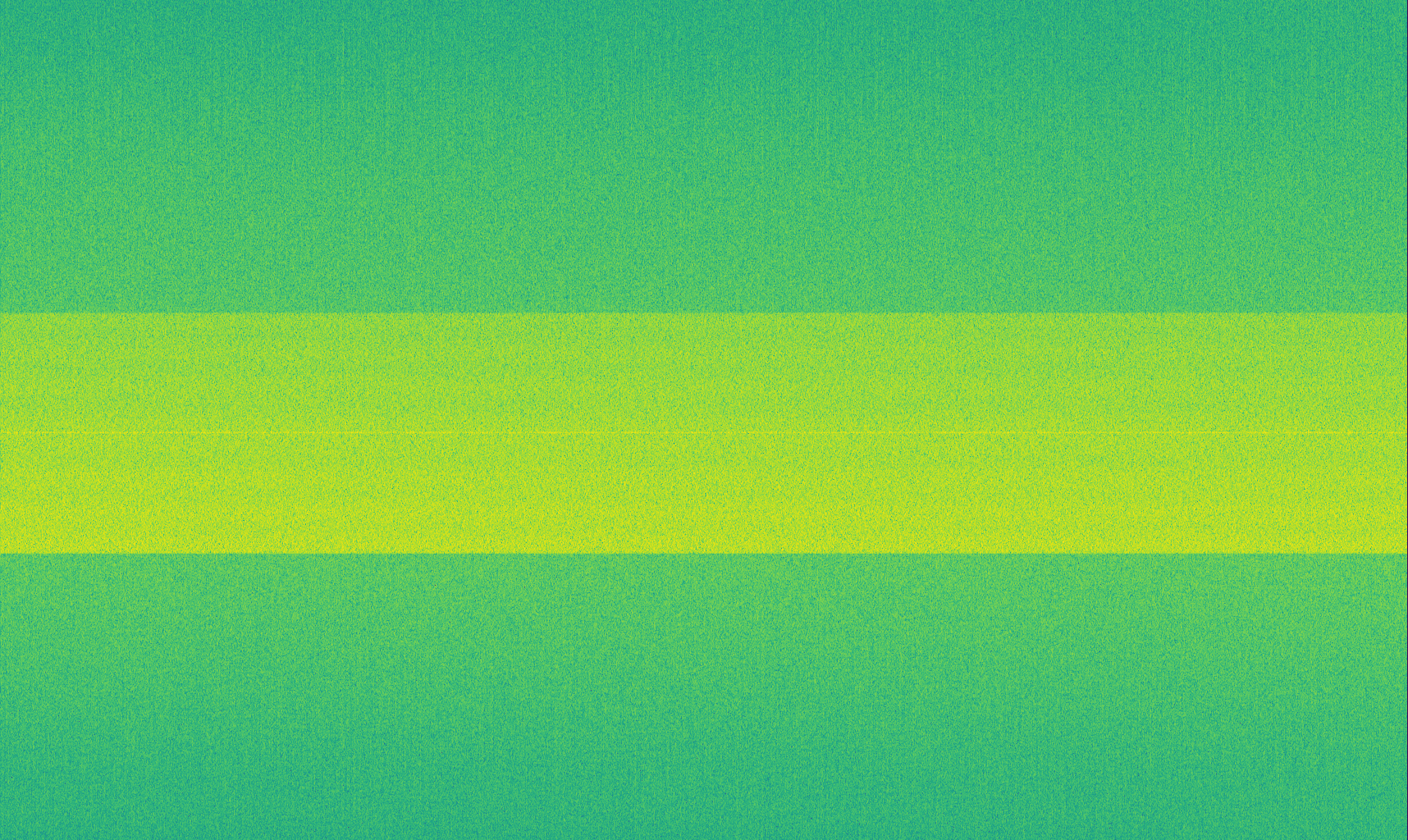}
        \caption{Autel X-Star}
    \end{subfigure}
    \begin{subfigure}[h]{0.45\columnwidth}
        \includegraphics[width=\columnwidth]{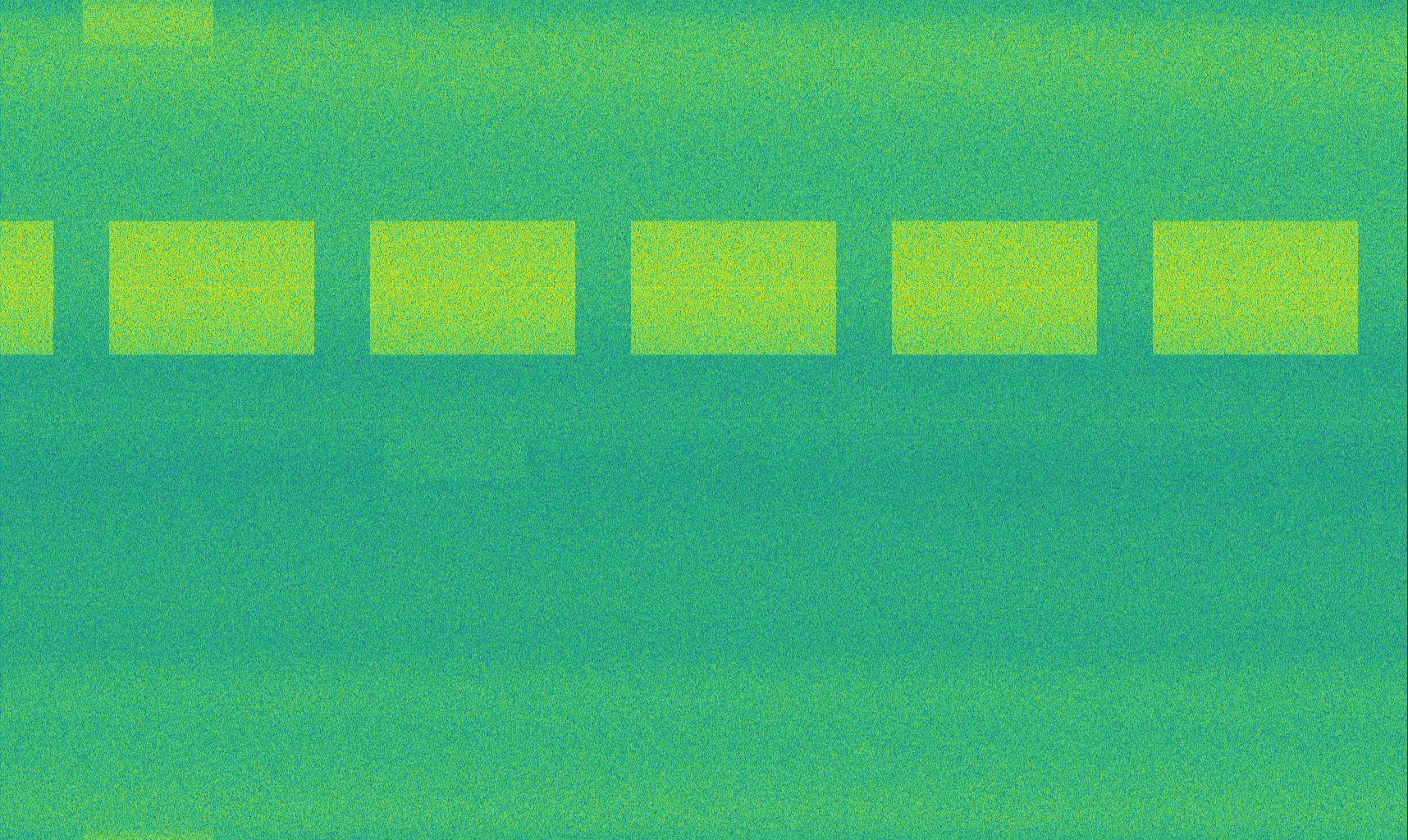}
        \caption{Autel EXOII}
    \end{subfigure}
    \begin{subfigure}[h]{0.45\columnwidth}
        \includegraphics[width=\columnwidth]{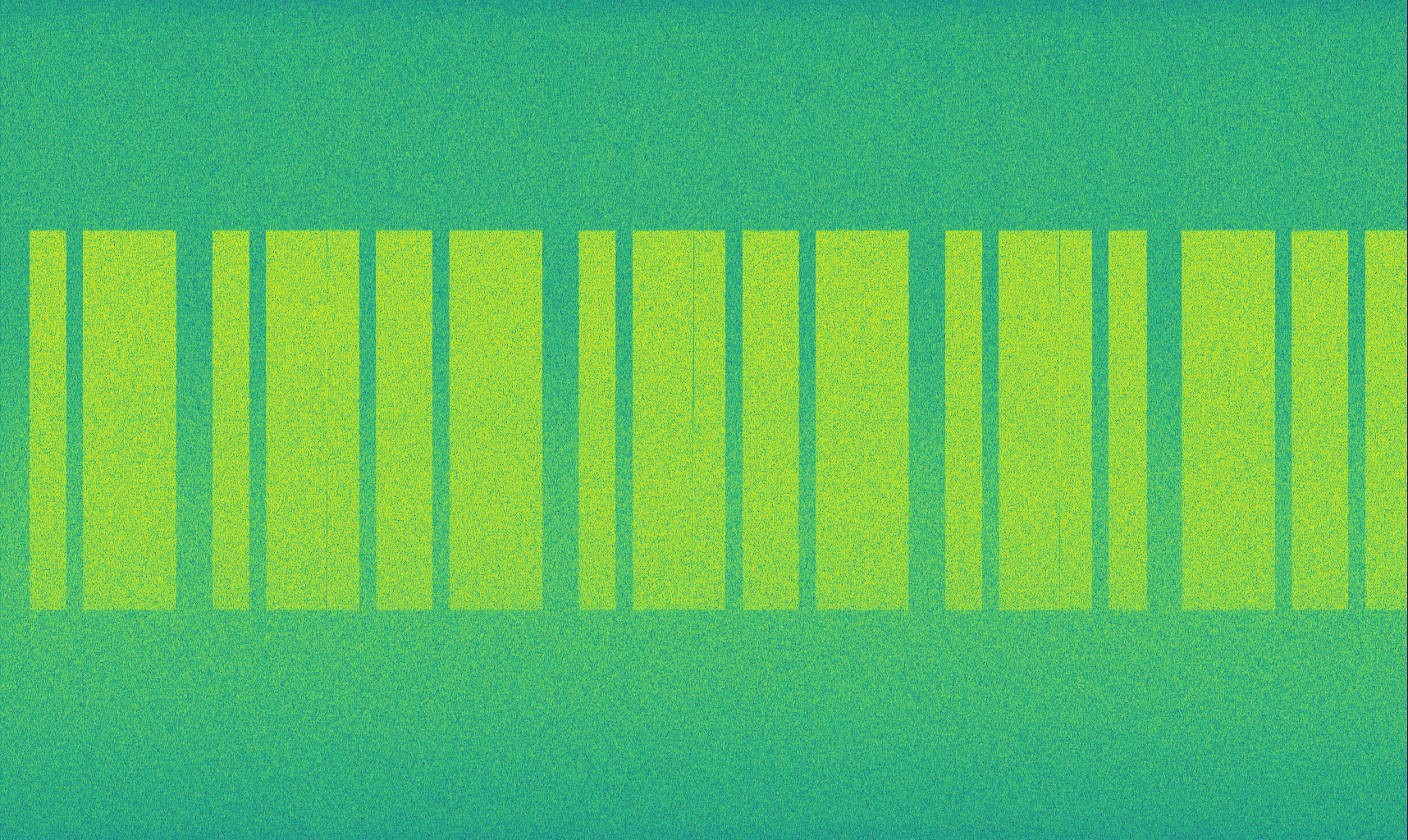}
        \caption{DJI Mavic 2 Pro}
    \end{subfigure}
    \begin{subfigure}[h]{0.45\columnwidth}
        \includegraphics[width=\columnwidth]{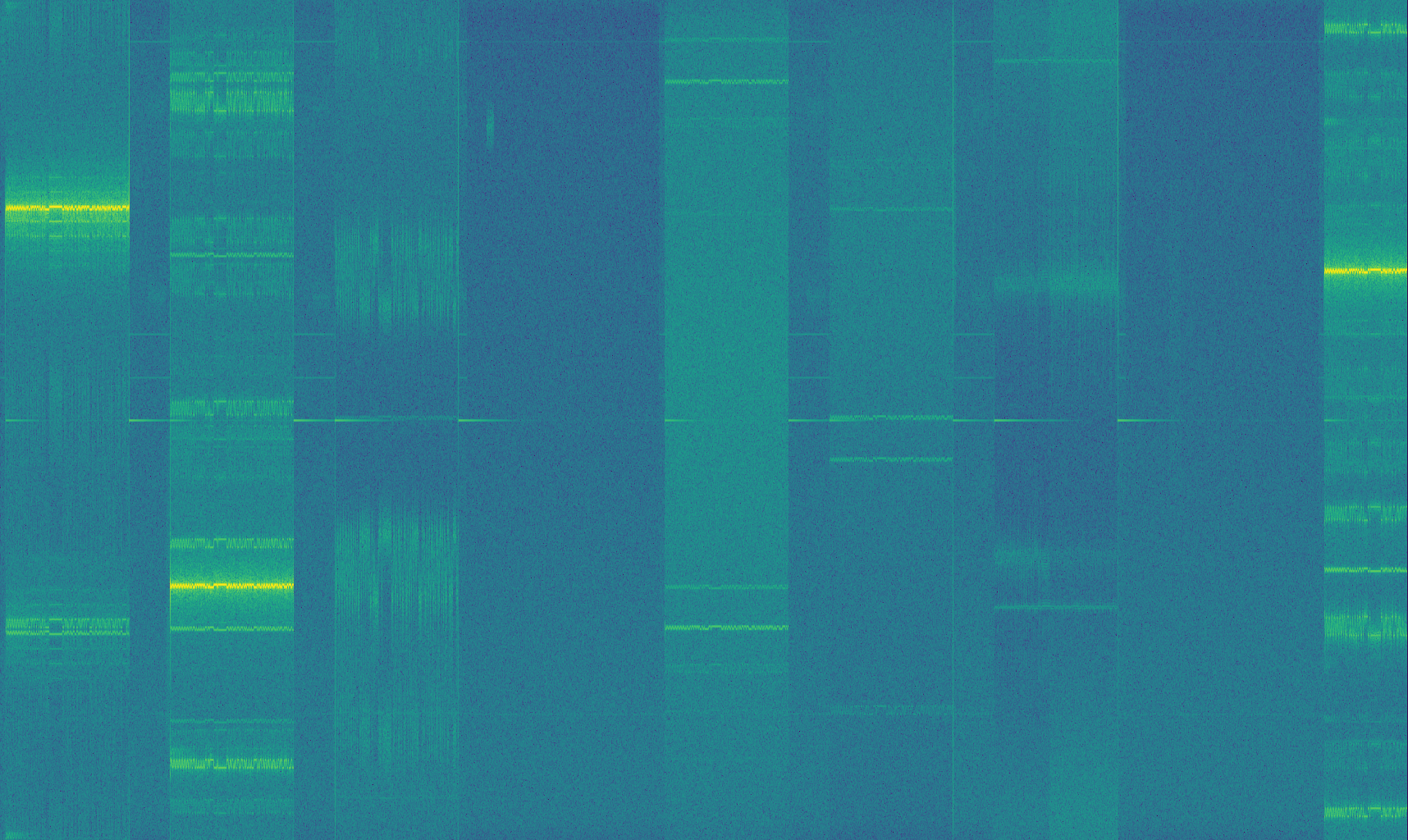}
        \caption{RadioMaster TX16S}
    \end{subfigure}
\caption{Representative spectrograms from the CDRF dataset, illustrating the diversity of signal types and device behaviors captured during controlled data collection.}
\label{fig:spec}
\end{figure}

{Next, we adjust the parameters on both the SDR device and the UAV to ensure proper alignment for optimal recording quality. This includes settings such as the SDR gain, center frequency, and the bandwidth of the UAV's video transmission. In this work, the SDR gain is set to 50\,dB for indoor use and 76\,dB for outdoor use, with a sampling rate of 20\,MHz in both environments. For the video signal from UAVs, we select one channel in each of the frequency bands supported by the UAV, e.g., 900\,MHz, 2.4\,GHz, or 5.8\,GHz. Generally, we choose channels that are not commonly occupied by non-UAV wireless traffic to ensure the quality of the collected data. Once the UAV transmission channel is defined, we tune the SDR center frequency to align with the UAV channel. This ensures that any UAV signal with a bandwidth of 20\,MHz or less is fully captured. Moreover, this frequency alignment simplifies the data preprocessing step for frequency-domain representations, because the resulting spectrogram contains a centralized UAV signal.}

{After that, we run the data recording script from the laptop for a predefined duration, and the raw data are stored in a \texttt{\{drone\_\allowbreak manufacturer\}\_\allowbreak\{drone\_\allowbreak model\}\_\allowbreak \{bandwidth\}\_\allowbreak\{drone\_\allowbreak center\_\allowbreak freq\}\allowbreak\_\allowbreak\{drone\_\allowbreak operation\_\allowbreak mode\}.dat} file along with all the predefined labels. In CDRF, we record the UAV manufacturer, model, center frequency of UAV signals, and bandwidth for each recording in the corresponding filename or directory name.} 

{Finally, we implement additional software scripts, including a signal playback module in GNU Radio, to load the raw data file, review its validity, and verify that the recorded signals match the predefined parameters. We also estimate the SNR and channel conditions to confirm that the recordings are in a sufficiently clean state. Fig.~\ref{fig:spec} shows representative spectrograms from the CDRF dataset, illustrating device behaviors captured during controlled data collection.}

\begin{figure}[ht]
    \centering
    \includegraphics[width=\columnwidth, trim={0 0 0 1.2cm}, clip]{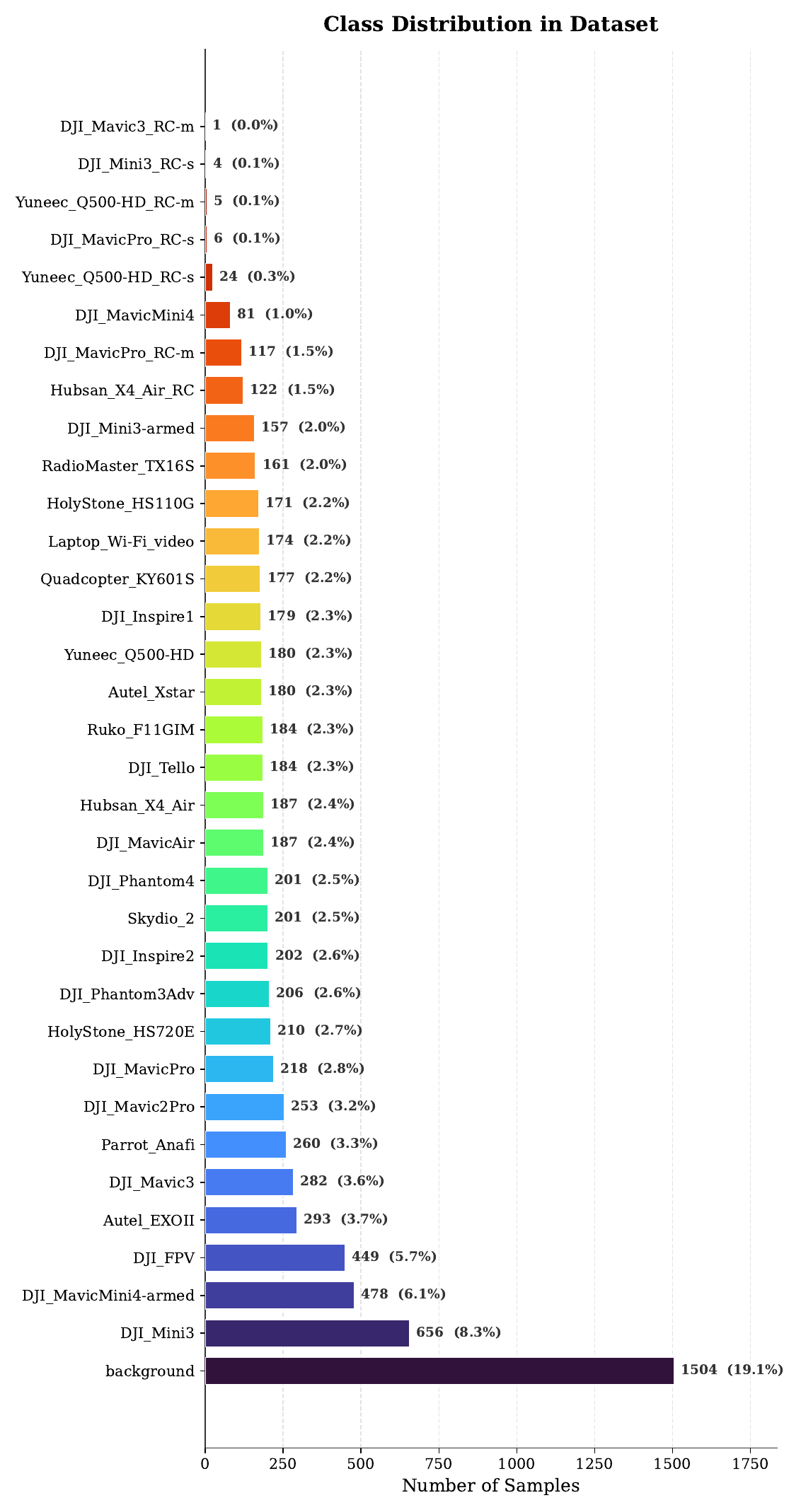}
    \caption{{Per-class sample distribution for the cage (indoor) subset of CDRF.}}
    \label{fig:dist_cage}
\end{figure}


\subsection{Collection of Outdoor Data}

{To complement our controlled indoor recordings, we conducted extensive outdoor data collection sessions aimed at enhancing the diversity and realism of the CDRF dataset by capturing UAV signals in complex RF environments that better reflect real-world operational conditions.}

{For each outdoor session, we recorded a comprehensive set of metadata to ensure reproducibility and utility. The raw I/Q data files are organized into a structured directory hierarchy where each directory name systematically encodes the key recording parameters. The naming convention is:}

\texttt{\{device\}\_\allowbreak\{status\}\_\allowbreak\{env\}\_\allowbreak\{sdr\_gain\}\_\allowbreak\{splitter\}\_\allowbreak\{duration\_recording\}\_\allowbreak\{distance\}\_\allowbreak\{altitude\}\_\allowbreak\{center\_freq\}\_\allowbreak\{drone\_c\_freq\}\_\allowbreak\{bw\}\_\allowbreak\{snr\}\_\allowbreak\{sampling\_rate\}\_\allowbreak\{record\_dir\}.dat}

The fields in the directory name are defined as:
\begin{itemize}
    \item \texttt{device}: The model of the drone or device being recorded.
    \item \texttt{status}: The operational state of the drone (e.g., hovering, flying, on the ground).
    \item \texttt{env}: The environment where the recording took place.
    \item \texttt{sdr\_gain}: The receiver gain of the SDR in dB.
    \item \texttt{splitter}: A flag indicating if a signal splitter was used.
    \item \texttt{duration\_recording}: The total duration of the recording in seconds.
    \item \texttt{distance}: The horizontal distance from the receiver to the drone in meters.
    \item \texttt{altitude}: The altitude of the drone in meters.
    \item \texttt{center\_freq}: The center frequency of the SDR receiver in MHz.
    \item \texttt{drone\_c\_freq}: The drone's transmission center frequency in MHz.
    \item \texttt{bw}: The signal bandwidth in MHz.
    \item \texttt{snr}: The estimated SNR in dB.
    \item \texttt{sampling\_rate}: The sampling rate of the SDR in MHz.
    \item \texttt{record\_dir}: The name of the directory where the recording is stored.
\end{itemize}

\begin{figure}[!t]
    \centering
    \begin{subfigure}[h]{0.45\columnwidth}
        \includegraphics[width=\columnwidth, trim={2.2cm 1.5cm 1.8cm 1.5cm}, clip]{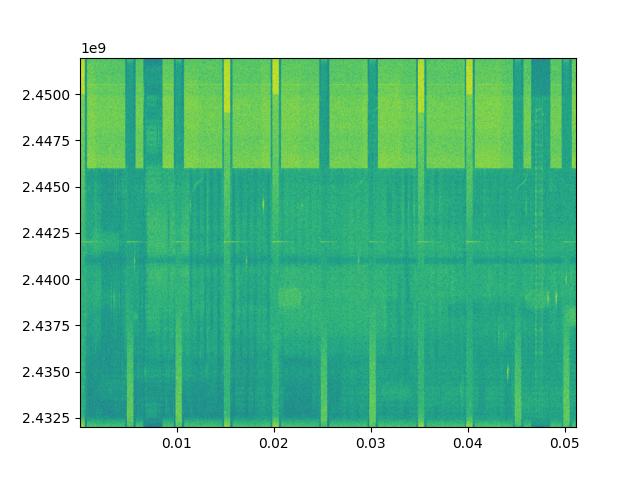}
        \caption{Autel EXOII}
    \end{subfigure}
    \begin{subfigure}[h]{0.45\columnwidth}
        \includegraphics[width=\columnwidth, trim={2.2cm 1.5cm 1.8cm 1.5cm}, clip]{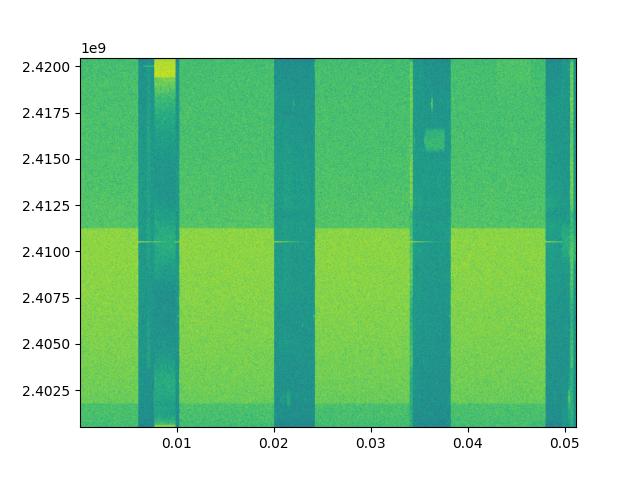}
        \caption{DJI Inspire1}
    \end{subfigure}
    \begin{subfigure}[h]{0.45\columnwidth}
        \includegraphics[width=\columnwidth, trim={2.2cm 1.5cm 1.8cm 1.5cm}, clip]{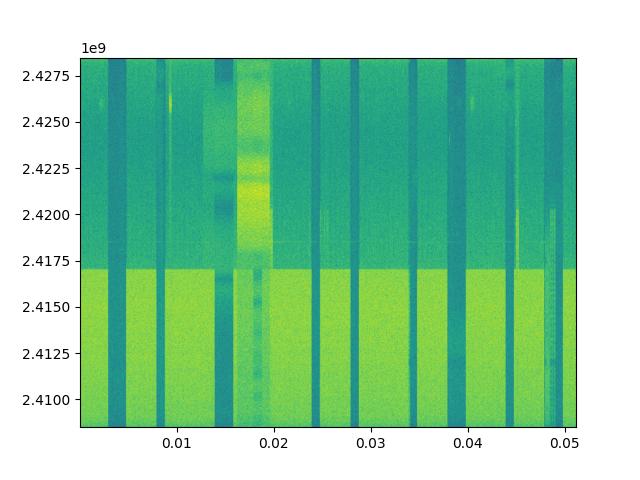}
        \caption{DJI Mavic3}
    \end{subfigure}
    \begin{subfigure}[h]{0.45\columnwidth}
        \includegraphics[width=\columnwidth, trim={2.2cm 1.5cm 1.8cm 1.5cm}, clip]{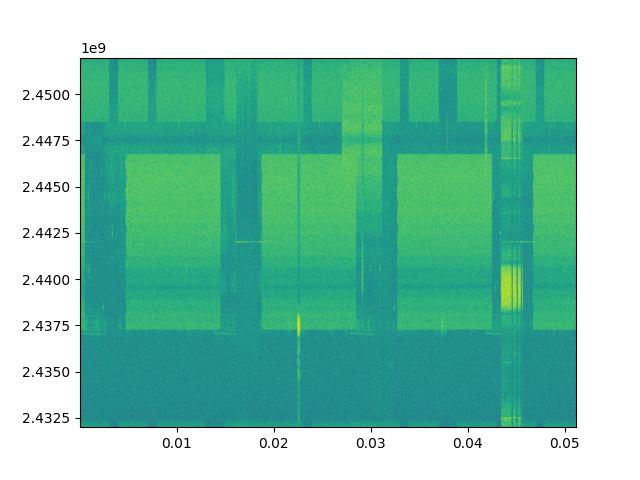}
        \caption{DJI Phantom3 Advanced}
    \end{subfigure}
\caption{Representative spectrograms from the outdoor dataset, showcasing the variability in signal characteristics due to environmental factors.}
\label{fig:outdoor_spec}
\end{figure}

{Fig.~\ref{fig:outdoor_spec} presents representative spectrograms from our outdoor data collection, featuring signals from the Autel EXOII, DJI Inspire1, DJI Mavic3, and DJI Phantom3 Advanced. In contrast to the clean, isolated signals captured indoors (Fig.~\ref{fig:spec}), these outdoor spectrograms exhibit significantly higher levels of background noise and interference, visually evident from the brighter and more cluttered backgrounds, a direct consequence of the complex real-world RF environments. The inclusion of such data is essential for developing and validating drone detection models robust enough for practical deployment.}

\begin{figure}[ht]
    \centering
    \includegraphics[width=\columnwidth, trim={0 0 0 1.2cm}, clip]{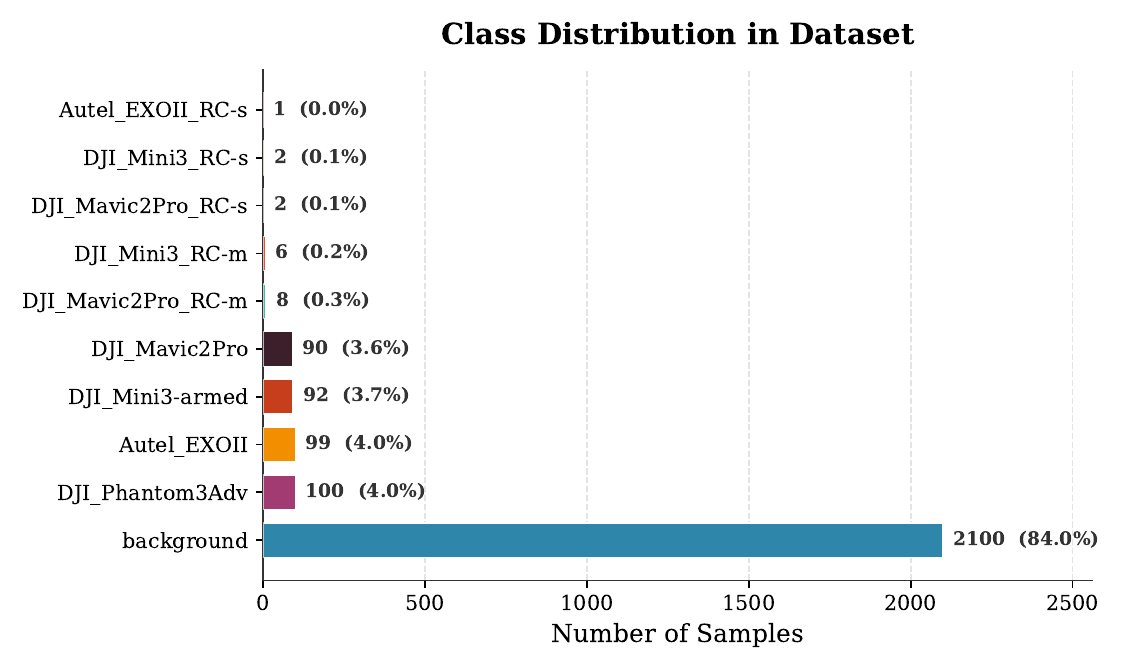}
    \caption{{Per-class sample distribution for the Rowan (outdoor) subset of CDRF.}}
    \label{fig:dist_rowan}
\end{figure}

\subsection{Dataset Composition and Signal Characteristics}
{The CDRF dataset provides a comprehensive and diverse collection of RF signals for drone detection and identification research, totaling approximately 500+\,GB of raw data.}

{At the core of the dataset is significant diversity in drone models and signal types. The collection includes signals from 39~unique classes, encompassing 23~distinct commercial and hobbyist UAV models, their remote controllers, and operational variants (e.g., armed vs.\ unarmed states). This variety, spanning major manufacturers such as DJI, Autel, and Parrot, ensures that models trained on CDRF are exposed to a wide range of communication protocols, representing a substantial leap from the handful of models found in many previous datasets. In addition to drone signals, we include a substantial number of no-drone recordings, capturing ambient RF noise and other common interferers such as Wi-Fi signals. This enables the training of robust binary classifiers that can distinguish between the presence and absence of drone activity, a critical requirement for real-world deployment. To further challenge detection and identification algorithms, the dataset also contains recordings of multiple drones operating simultaneously.}

{All raw data are stored using the In-phase and Quadrature (I/Q) sampling method. Each sample is represented as an interleaved binary bitstream of two 32-bit floating-point numbers, capturing the full complex signal. This raw format is essential, as it allows researchers to develop and test novel signal processing and feature extraction techniques that operate directly on the baseband signal prior to any transformation such as the STFT.}

{The signals captured in CDRF exhibit a variety of characteristics typical of modern drone communication systems. Many drones utilize Frequency-Hopping Spread Spectrum (FHSS) for their command and control links. These signals are characterized by key parameters such as hopping frequency, duration, duty cycle, and hopping-pattern period. In contrast, signals used for video transmission typically have wider bandwidth and longer duration than control signals. We refer to the collection of these characteristics as the \emph{RF Drone Fingerprint}, which serves as a rich input for deep learning models to differentiate between drone types.}

{Furthermore, the data capture the inherent behavioral variability of drone signals. For instance, the frequency-hopping patterns of a drone during its initial pairing process can differ significantly from those during normal flight operations. The dataset also captures more subtle phenomena, such as the variable duration of video transmission signals from certain drone models, a characteristic that can serve as an additional feature for robust classification.}

\begin{figure}[ht]
    \centering
    \includegraphics[width=\columnwidth, trim={0 0 0 1.2cm}, clip]{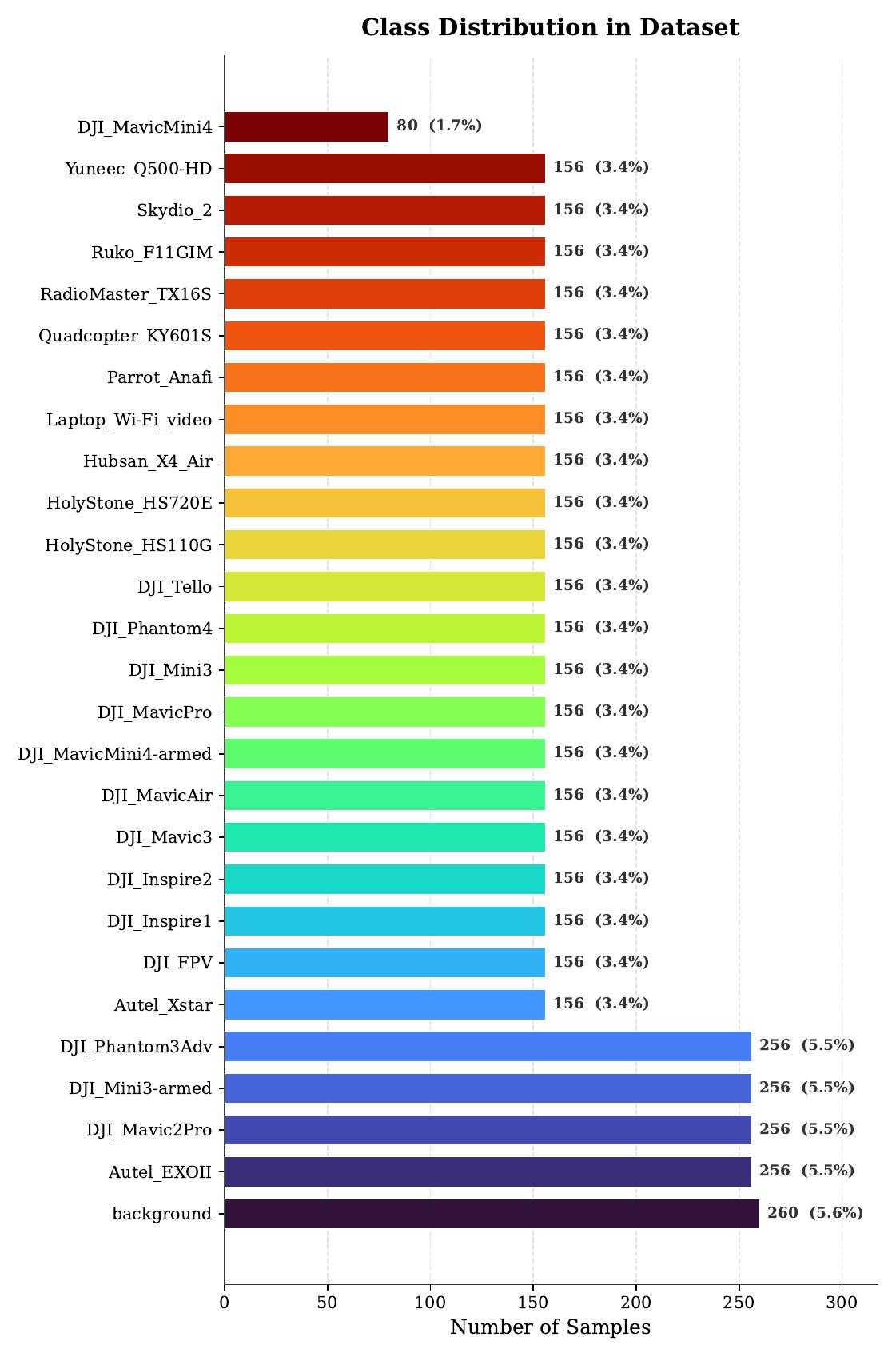}
    \caption{{Per-class sample distribution for the final balanced dataset of CDRF.}}
    \label{fig:dist_balanced}
\end{figure}

{To provide a quantitative overview of the dataset composition, we present the per-class sample distributions for the three primary subsets of CDRF. Fig.~\ref{fig:dist_cage} shows the distribution of samples collected in the controlled Faraday-cage environment, which provides clean, high-SNR reference captures. Fig.~\ref{fig:dist_rowan} shows the distribution of outdoor recordings collected at the Rowan University campus, reflecting real-world interference and environmental variability. Fig.~\ref{fig:dist_balanced} presents the final balanced dataset used for training and evaluation, illustrating the class-balancing strategy applied to mitigate skewed learning.}

\section{Data Preprocessing, Augmentation, and Annotation}
\label{sec:data_processing}

We release both a raw-signal processing pipeline and augmentation utilities designed to create realistic, label-consistent training data from complex I/Q captures. All components are open-source and parameterized for reproducibility.

\subsection{Spectrogram Generation}
Raw complex I/Q streams are transformed into time–frequency representations via the STFT. Given a complex signal $x[n]$, window $w[n]$, window length $N$, and hop size $H$, the STFT is ~\cite{oppenheim1999discrete}
\[
X(m,k) \;=\; \sum_{n=0}^{N-1} x[n+mH]\; w[n]\; e^{-j 2\pi kn/N},
\]
where $m$ is the discrete time index and $k$ is the discrete frequency index.

We compute the power spectrogram $S(m,k) = \lvert X(m,k)\rvert^2$ and render in dB as
\[
S_{\mathrm{dB}}(m,k) \;=\; 10\log_{10}\big(S(m,k)+\epsilon\big),
\]
{where $\epsilon$ is a small positive constant used to avoid taking the logarithm of zero, followed by mapping to RGB using a perceptual colormap. By default, we use \texttt{scipy.signal.stft}~\cite{virtanen2020scipy} with a Hann window, $N=\texttt{FFT\_SIZE}=1024$, overlap $=128$ samples, and return two-sided spectra with DC centered via \texttt{fftshift}. Default sampling is $\texttt{SAMPLING\_RATE}=20$\,MHz and we generate $\texttt{NUM\_FFT\_SPEC}=1500$ time bins per sample. Spectrograms are exported as PNG images alongside per-sample metadata.}

\subsection{Frequency Resolution and Colormap Selection}
The window size $N$ and hop $(N-\text{overlap})$ set the time–frequency trade-off; larger $N$ improves frequency resolution at the cost of temporal precision. All parameters are exposed to support ablations. We provide multiple colormaps (\texttt{viridis}, \texttt{plasma}, \texttt{inferno}, \texttt{magma}, \texttt{cividis}, \texttt{gray}, \texttt{hot}) to test color sensitivity; the default pipeline normalizes spectrograms to $[0,1]$ before colorization and composites to RGB for consistent rendering.

\subsection{Data Normalization}
We include signal-level and spectrogram-level normalization operators:
\begin{itemize}
    \item \emph{Power normalization} of complex I/Q to unit average power: $\tilde{x}=x/\sqrt{\mathbb{E}\lvert x\rvert^2+\epsilon}$.
    \item \emph{Z-normalization} of real and imaginary parts independently to zero mean, unit variance.
    \item \emph{Spectrogram normalization} options: per-frequency z-score, per-time z-score, or global min–max to $[0,1]$ prior to colorization.
\end{itemize}

These are selectable in the augmentation scripts and can be toggled per experiment.

\begin{figure}[!t]
    \centering
    \begin{subfigure}[h]{\columnwidth}
        \centering
        \includegraphics[width=\columnwidth]{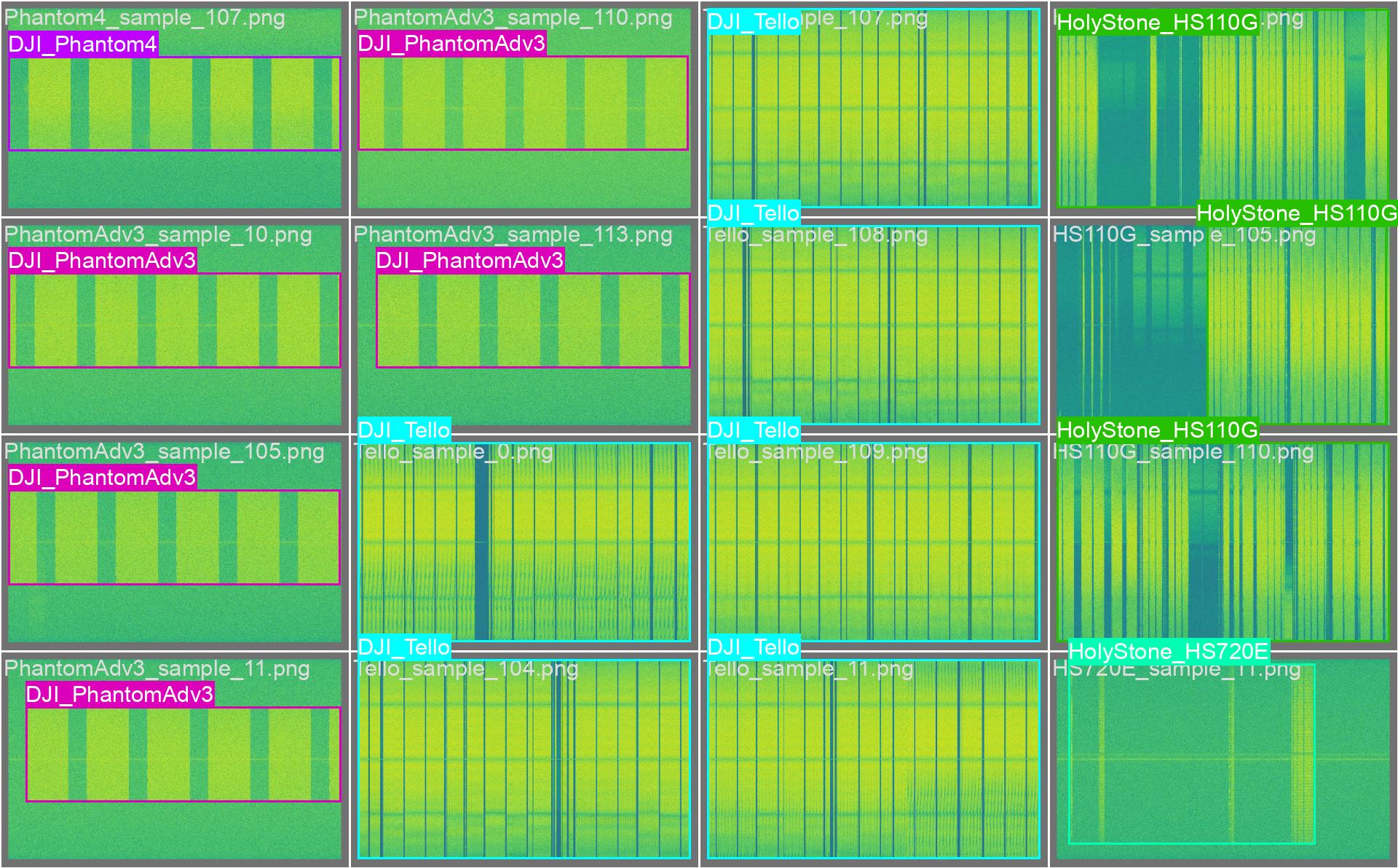}
        \caption{Annotated Data.}
    \end{subfigure}
    \hfill
    \begin{subfigure}[h]{\columnwidth}
        \centering
        \includegraphics[width=\columnwidth]{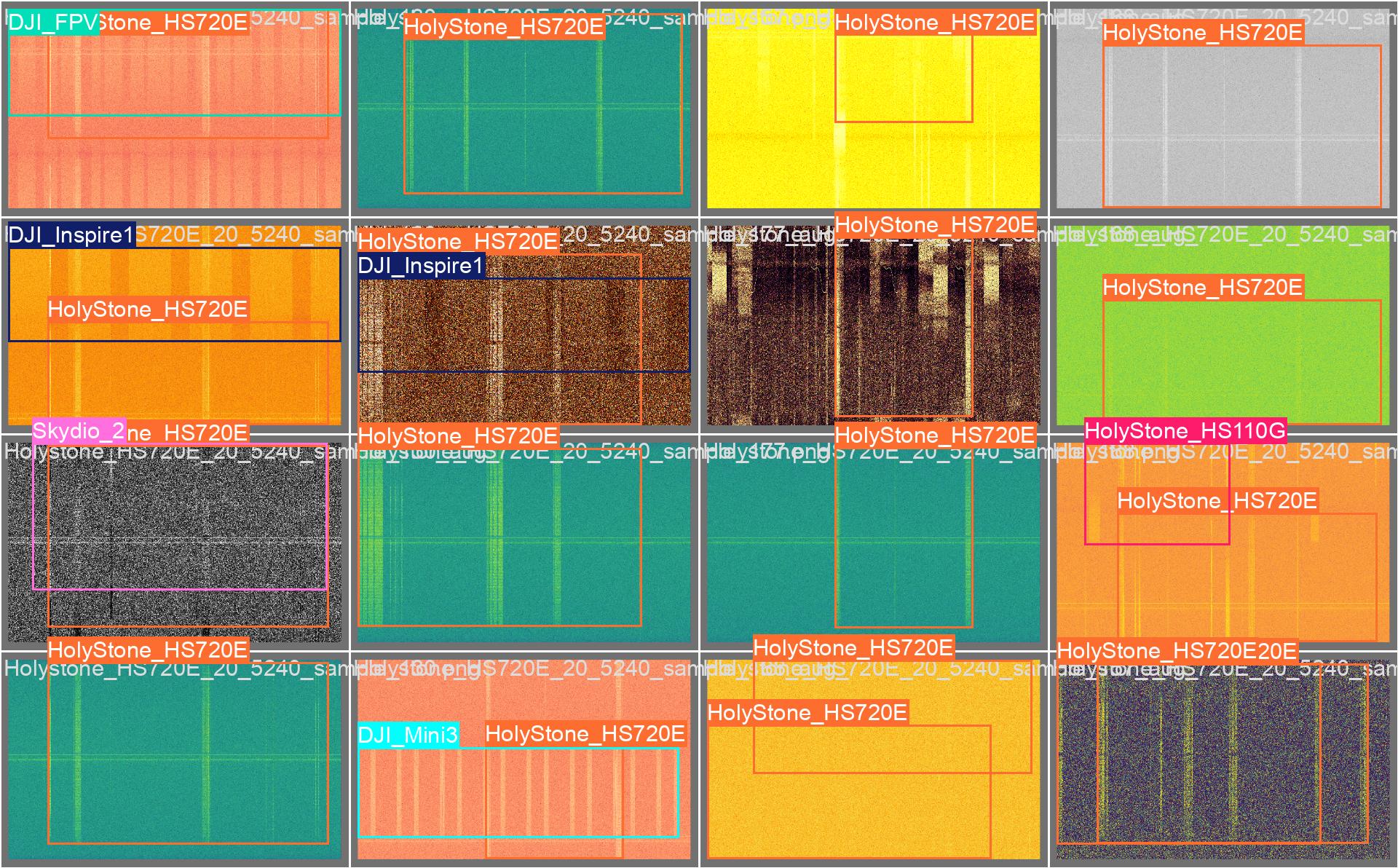}
        \caption{Augmented Data.}
    \end{subfigure}
    \caption{Examples of annotated and augmented data.}
    \label{fig:augmented_data}
\end{figure}

\subsection{Custom Augmentation Tools}
Our augmentation operates at the \emph{raw-signal} (I/Q) level to better preserve RF physics and then projects to spectrograms, with exact label propagation to detection annotations.

\subsubsection{SNR Conditioning via AWGN}
To synthesize controlled SNRs, we add complex AWGN $n\sim\mathcal{CN}(0,\sigma^2)$ to $x$ where the noise power $\sigma^2$ is chosen for target $\mathrm{SNR}_{\mathrm{dB}}$:
\[
P_s=\mathbb{E}\lvert x\rvert^2,\quad \mathrm{SNR}_{\mathrm{lin}}=10^{\mathrm{SNR}_{\mathrm{dB}}/10},\quad\sigma^2=P_s/\mathrm{SNR}_{\mathrm{lin}}.
\]  

We optionally export noise-only spectrograms for analysis and create SNR-stratified dataset variants (e.g., per-SNR subdirectories), enabling robustness sweeps.

\subsubsection{Frequency Shifting}
We simulate carrier offsets and front-end errors by multiplying with a complex exponential
\[
x_{\Delta f}[n] \;=\; x[n]\; e^{j2\pi \Delta f \, n/F_s},
\]
which induces a vertical translation on the spectrogram and where $F_s$ is sampling rate. This is applied before STFT and is label-consistent with our detection annotation recomputation (below).

\subsubsection{Interferer Mixing}
To emulate spectrum crowding, we mix two normalized signals at a controllable ratio $\alpha\in[0,1]$:
\[
x_{\text{mix}} \;=\; \operatorname{norm}\big(x_1\;+\;\alpha\, x_2\big),
\]
optionally with independent frequency shifts on each source. This produces realistic co-channel scenarios directly in the I/Q domain.

\subsubsection{Spectrogram Rendering Variants}
Augmented samples are rendered under randomized, but reproducible, choices of colormap and normalization policy to increase visual diversity while keeping frequency content intact.

\subsubsection{Background Noise and Non-Target Signals}
Our dataset includes a comprehensive corpus of non-target signals, captured from diverse environments where no drones were active. To facilitate common ML workflows, we provide a suite of tools to extract these ``no-drone" segments and generate unified metadata files for mixed-signal experiments. This curated collection of real-world negative samples is crucial for training well-calibrated binary detectors and is explicitly designed to support advanced strategies like hard-negative mining.

\subsection{Annotation Strategy}
We adopt YOLO-format annotations on spectrogram images with a \emph{whole-signal} policy: contiguous RF emission bursts from a device are annotated as a single object to capture both spectral morphology and temporal occupancy. Fig.~\ref{fig:augmented_data} shows examples of annotated data and augmented data with annotations.

{In general, a spectrogram of a UAV signal segment consists of a sequence of ON--OFF states, which represents the transmission pattern of the UAV in both the time and frequency domains. To enable the YOLO model to better capture signal characteristics in both domains, we define the annotation policy such that the bounding box for each signal type in a spectrogram image starts and ends with an ON~state. OFF~states at the beginning and end of a spectrogram are excluded. This helps the YOLO model track the start and end of a transmission and thus produce more accurate bounding boxes. If the RF emission bursts captured by a spectrogram occupy less than 10\% of the spectrogram's time extent, or if there is only a single ON~state in the entire spectrogram, no bounding box annotation is assigned and the spectrogram is labeled as background.} 

{As for the RC signals, due to their frequency-hopping property, CDRF provides two types of YOLO-formatted annotations to support flexibility in future research and deployment:}
\begin{itemize}
    \item {A single bounding box covering all the RC signal bursts captured in a spectrogram. As with UAV signal labeling, the edges of the bounding box are aligned with the start and end of the RC burst sequence.}
    \item {Per-channel annotations for RC signal bursts transmitted on the same frequency channel. Because frequency-hopping behavior is not guaranteed to occur or be captured in every spectrogram within a continuous recording, this annotation type can improve YOLO learning and detection performance in real-world deployment.}
\end{itemize}

{Moreover, because most RC signals are narrowband and short in duration for each ON~state, we do not assign bounding box annotations to spectrograms containing only a single ON~state. A bounding box covering a single ON~state does not provide sufficient information for the model to learn the signal's pattern across either the time or frequency domain. A similar observation regarding UAV video signal annotation is reported in~\cite{nelega2023radio}. Furthermore, detection or classification based on a single ON~state is generally neither expected nor considered reliable in real-world UAV detector deployment scenarios.}

\subsubsection{Label-Preserving Transforms with Wrap-Around}
Frequency shifts correspond to vertical translations in normalized image coordinates by $\Delta y=\Delta f/F_s$. We recompute bounding boxes analytically after augmentation and correctly handle wrap-around at the top/bottom edges: if a box straddles the boundary after shift, it is split into two valid boxes whose heights sum to the original. Boxes are also filtered by a configurable minimum height to avoid degenerate labels after extreme shifts.

\subsubsection{Dataset Hygiene and Third-Party Integration}
We include utilities to normalize third-party releases (e.g., Roboflow-style layouts) into a consistent hierarchy (\texttt{images/}\,\,\texttt{labels/} with class folders and \{train,val,test\} splits) and to preserve class mappings. This enables direct augmentation and training across heterogeneous sources.

\subsection{Reproducibility and Metadata}
Each sample carries rich metadata to ensure reproducibility: original file path, device/model label, sampling rate, FFT size, number of STFT frames, per-sample time bounds (start/end in seconds), center frequency, and output paths for spectrograms. All processing parameters (e.g., window size, overlap, SNR target, shift amounts, mixing weights) are saved or deterministically re-creatable from configuration, facilitating exact regeneration of training splits and ablation studies.

%% file: sections/4_experiments_results.tex
\section{Machine Learning Tasks and Baselines}
\label{sec:ml_tasks}
{This section details the ML tasks supported by CDRF, the baseline models employed, and their performance. We address drone detection, single-label classification, hierarchical classification, and open-set recognition.}

\subsection{{Baseline Model Selection Rationale}}
\label{subsec:model_selection}

{The baseline models in this work are chosen to reflect an end-to-end, edge-deployable RF perception pipeline in which raw signal capture, spectrogram generation, and model inference must all execute within tight latency and memory budgets on resource-constrained hardware. For detection, we adopt YOLOv11n (nano), the smallest variant of the YOLOv11 family. In a practical deployment scenario, an edge device must continuously digitize the RF front-end output, compute the STFT to produce a spectrogram, and run inference before the next observation window arrives; a larger backbone (e.g., YOLOv11l/x or a two-stage detector) would dominate this budget and preclude real-time operation. YOLOv11n therefore represents a realistic operating point rather than a pursuit of maximum accuracy.}

{For classification and hierarchical tasks we use ResNet-18, a compact yet well-characterized architecture that has been widely adopted as a standard baseline in RF and spectrogram classification literature. Together, these lightweight models establish reproducible performance floors against which the community can benchmark heavier or more specialized architectures on the CDRF dataset.}

\subsection{Drone Detection (YOLO)}
\label{subsec:yolo_experiments}

\subsubsection{Task Definition}
{Given a time-frequency spectrogram, the goal is to detect and localize RF emissions attributable to drones. The detector predicts a set of axis-aligned bounding boxes with class labels in YOLO normalized coordinates, where each box corresponds to a contiguous emission burst (whole-signal policy) on the spectrogram.}

\subsubsection{Baseline Model Architecture and Training}
{We use a single-stage object detector, \texttt{YOLOv11n}~\cite{khanam2024yolov11} (Ultralytics~\cite{jocher2022ultralytics} v8.3.156), trained on spectrograms produced by our pipeline (\cref{sec:data_processing}). We evaluate two scenarios: (i)~Clean, in which spectrograms are generated without augmentation; and (ii)~Augmented, in which raw complex I/Q signals are transformed before the STFT via frequency shifting, interferer mixing, and SNR conditioning. Models are initialized from COCO-pretrained weights~\cite{lin2014microsoft,jocher2022ultralytics,khanam2024yolov11}, trained with deterministic seeds, and validated at each epoch. Labels follow the whole-signal policy, and for frequency shifts we recompute YOLO annotations exactly with wrap-around handling on the frequency axis to preserve label consistency. Unless otherwise specified, Ultralytics defaults are used for input size, data augmentation, and non-maximum suppression. {The clean model is trained for 100~epochs, whereas the augmented model is trained for 50~epochs. This discrepancy is deliberate: because the augmented dataset is constructed by applying multiple transformations (frequency shifts, interferer mixing, SNR conditioning) to the original images, each source sample is effectively seen several times per epoch through its augmented variants. Training for fewer epochs therefore yields a comparable effective exposure to the underlying data while avoiding overexposure that could bias the comparison. All reported results correspond to the best validation checkpoint selected by monitoring mAP@[.5:.95] across epochs.}}

\subsubsection{Evaluation Metrics}
{We report standard detection metrics: class-aggregated Precision and Recall; mean Average Precision at IoU~0.5 (mAP@0.5); and COCO-style mAP averaged over IoU thresholds from 0.5 to 0.95 in steps of 0.05 (mAP@[.5:.95]). For qualitative error analysis, we include normalized confusion matrices computed from matched detections.}

\subsubsection{Experimental Results and Baseline Performance}
\begin{table}[hht]
\centering
\caption{YOLO detection results.}
\label{tab:yolo_detection_results}
\resizebox{\columnwidth}{!}{%
\begin{tabular}{lcccccc}
\toprule
Run & Epochs & Precision & Recall & mAP@0.5 & mAP@[.5:.95] \\
\midrule
Clean & 100 & 0.932 & 0.982 & 0.986 & 0.948\\
Augmented & 50 & 0.940 & 0.855 & 0.935
& 0.838\\
\bottomrule
\end{tabular}%
}
\end{table}

{Table~\ref{tab:yolo_detection_results} reports the best validation checkpoint for each training regime, selected by the highest mAP@[.5:.95] observed during training.} The model trained without any augmentations (referred to as ``clean") achieves strong performance (Precision $=0.932$, Recall $=0.982$, mAP@0.5 $=0.986$, mAP@[.5:.95] $=0.948$), with the top mAP@[.5:.95] observed near epoch~78. Under augmentation, precision is maintained and slightly improved (Precision $=0.940$) while recall drops (Recall $=0.855$), yielding lower top‑line mAP (mAP@0.5 $=0.935$, mAP@[.5:.95] $=0.838$; best near epoch~49). This reflects the expected precision-recall trade-off when training on harder, low-SNR and interferer-rich conditions: the detector becomes more conservative (fewer false positives) at the expense of increased misses (more false negatives).

\begin{figure}[htt]
    \centering
    \begin{subfigure}[h]{\columnwidth}
        \centering
        \includegraphics[width=\columnwidth]{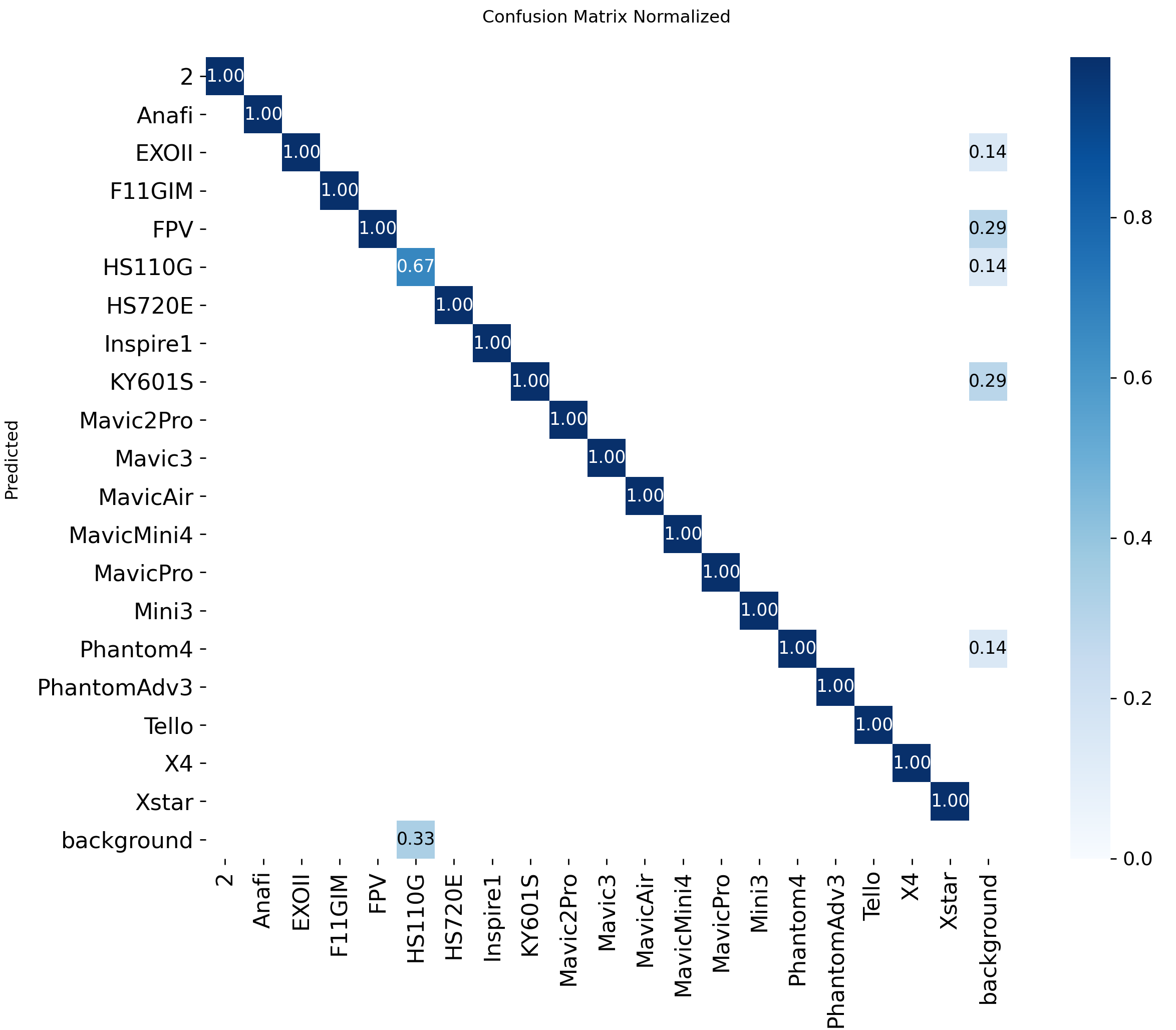}
        \caption{Clean: normalized confusion.}
    \end{subfigure}
    \hfill
    \begin{subfigure}[h]{\columnwidth}
        \centering
        \includegraphics[width=\columnwidth]{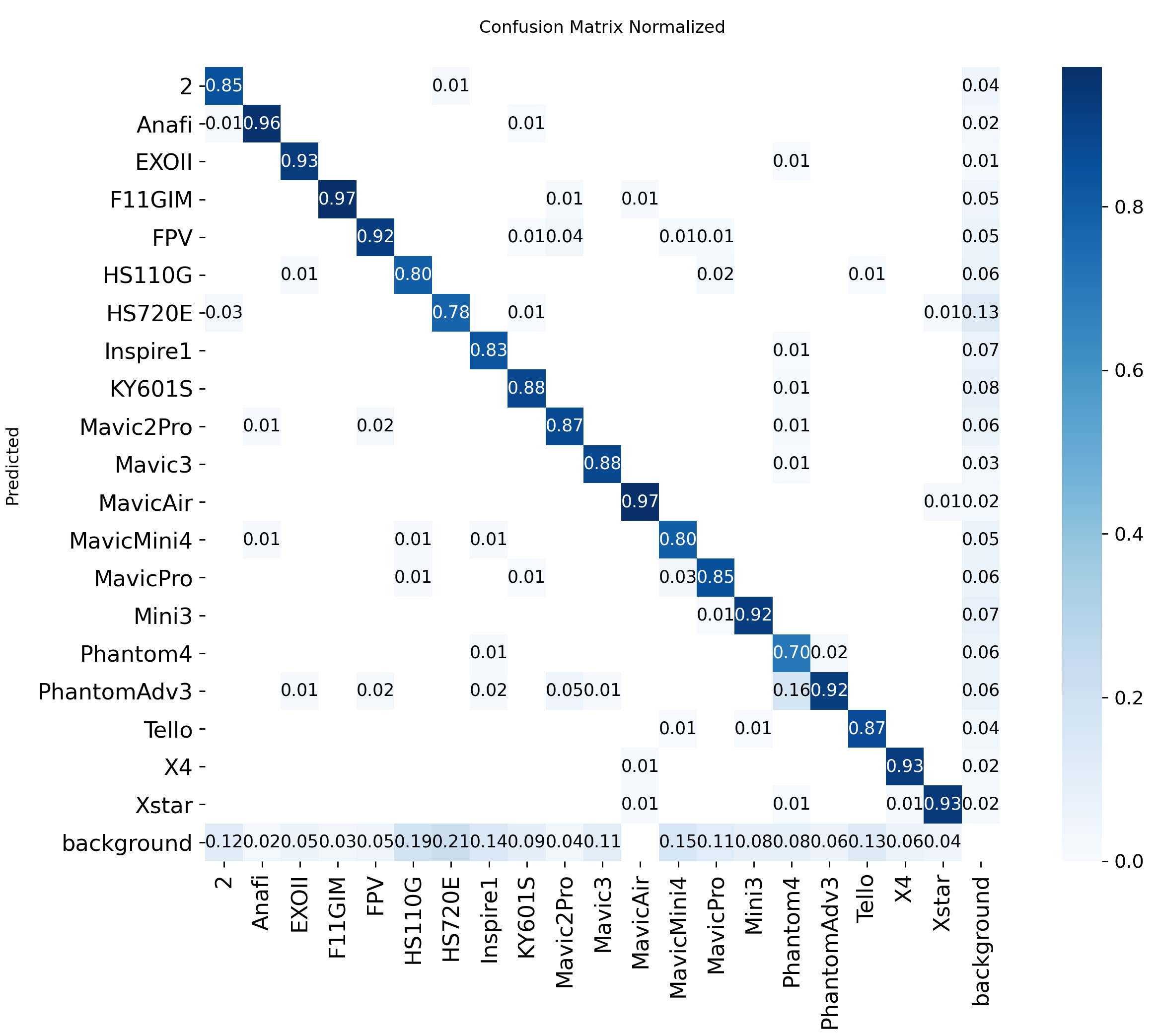}
        \caption{Augmented: normalized confusion.}
    \end{subfigure}
    \caption{Class-wise performance and error structure on validation.}
    \label{fig:yolo_confusions}
\end{figure}

{Despite this shift, the normalized confusion matrices in Fig.~\ref{fig:yolo_confusions} remain strongly diagonal for both regimes, indicating that whole-signal annotations preserve class separability even under frequency shifts and spectrum crowding. The dominant degradation under augmentation is recall (missed detections) rather than systematic label confusion, suggesting room for robustness gains via longer training, class-balanced sampling, or stronger backbones without altering the annotation policy.}

\subsection{Hierarchical Classification}
\subsubsection{Task Definition}
\begin{figure*}[phtb] 
    \centering
    \includegraphics[width=1\linewidth]{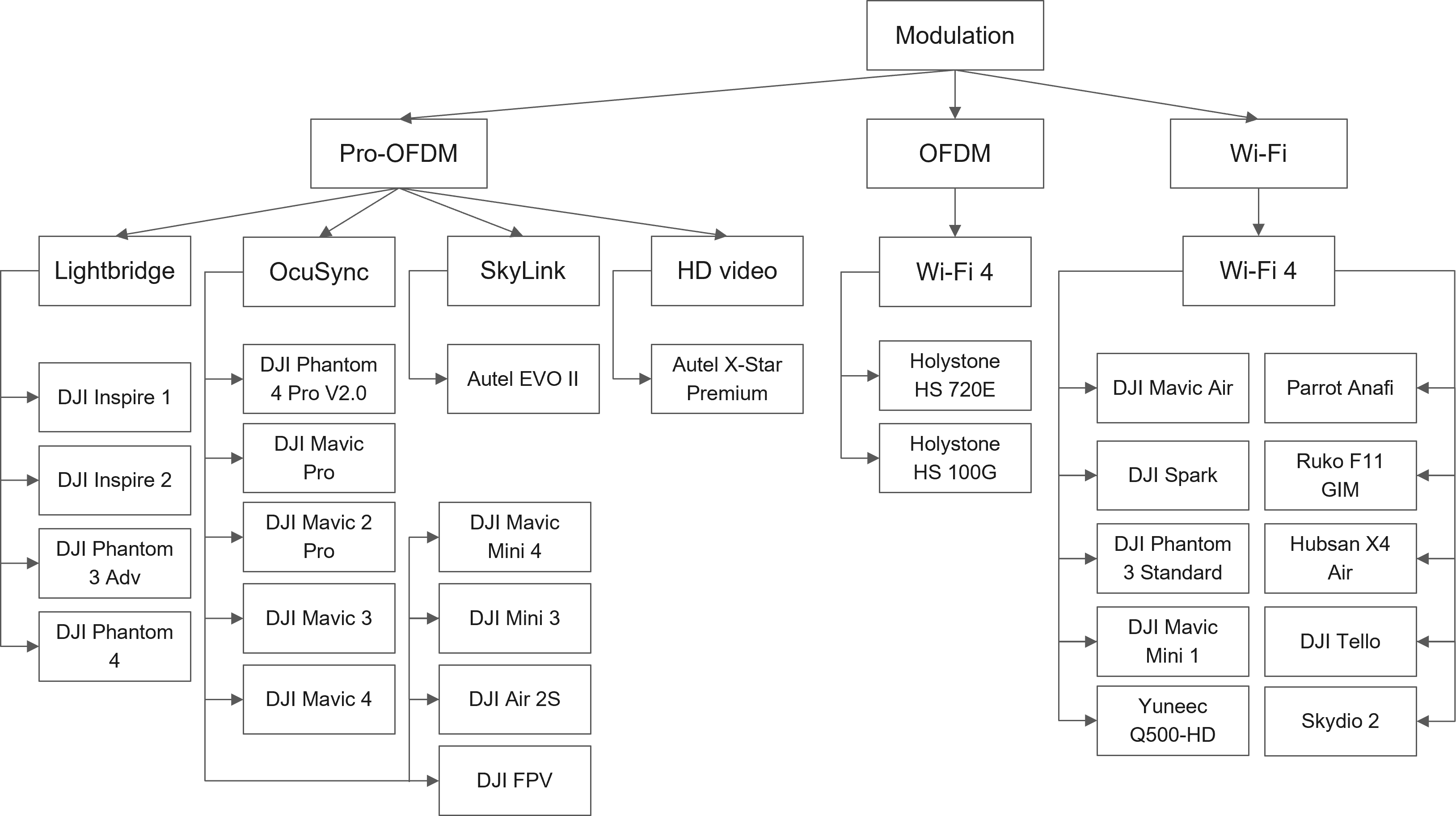}
    \caption{Hierarchical classification tree structure of CageDroneRF.}
    \label{fig:hier}
\end{figure*}

{To provide more robust and interpretable classifications, we explore a hierarchical approach that classifies signals in a coarse-to-fine manner. Fig.~\ref{fig:hier} presents the full structural tree of the hierarchical classification. Our hierarchy is structured into three levels: Modulation (3~classes), Protocol (5~classes), and Model (27~classes).}

\subsubsection{Baseline Model Architecture and Training}
{We implement a multi-task learning model using a shared ResNet-18~\cite{he2016deep} backbone to extract features from input spectrograms. The output of the feature extractor is fed into three separate classification heads, each dedicated to one hierarchical level. The model is trained end-to-end using a HierarchicalLoss, which is a weighted sum of the cross-entropy losses from each head.}

{To evaluate the effectiveness of our indoor data and its generalization to outdoor scenarios, we train our hierarchical classifier on the indoor subset of CDRF and evaluate it on the outdoor subset. We further examine how including outdoor data in the training set affects performance by training a second model on the combined indoor and outdoor data. In both cases, the amount of indoor training data is kept constant to ensure a fair comparison. The models face challenging scenarios including varying SNR levels, signal bandwidths, different drone altitudes and distances, and the presence of various interferers.}

\subsubsection{Experimental Results and Baseline Performance}

\begin{figure}[phtb] 
    \centering
    \includegraphics[width=1\linewidth]{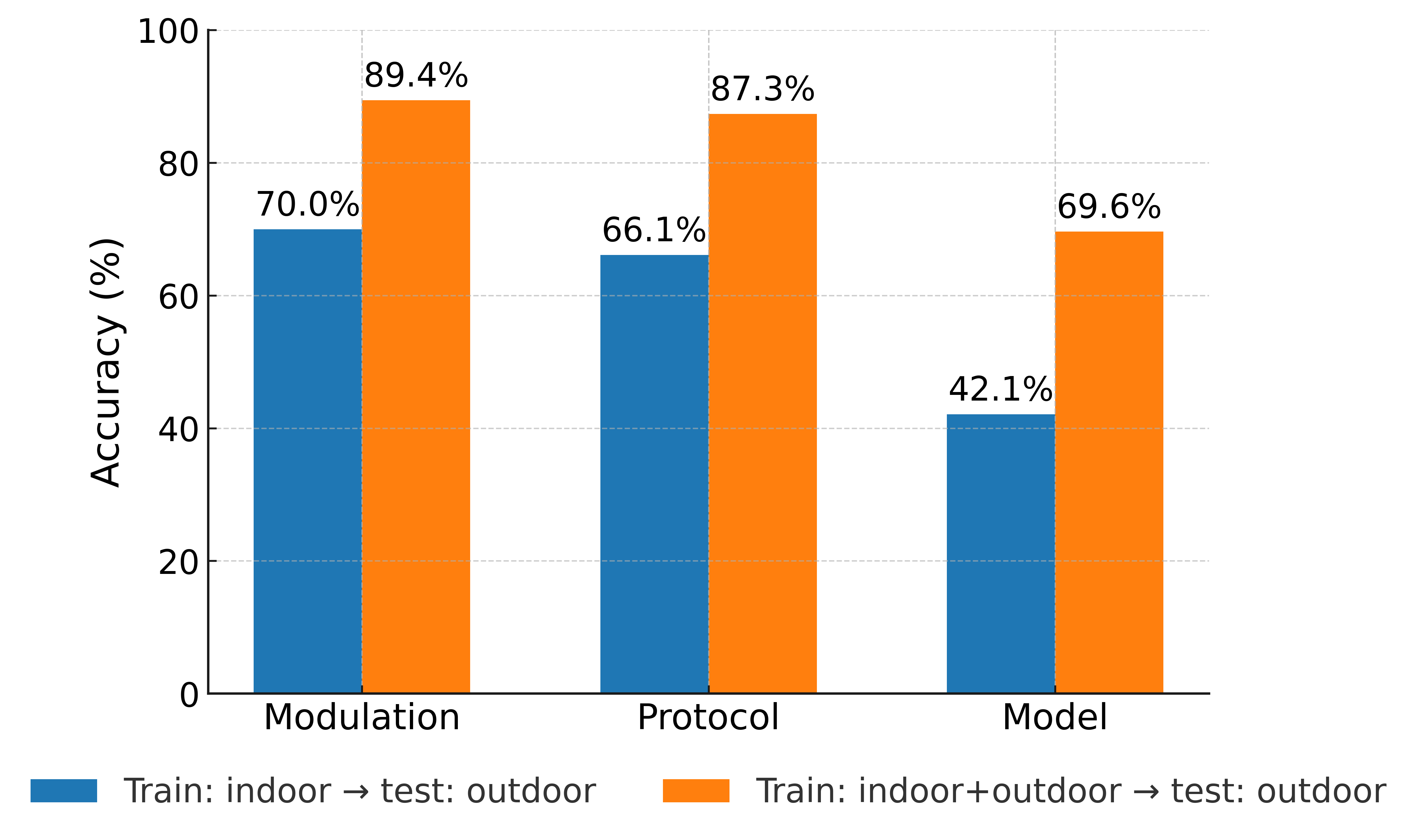}
    \caption{Hierarchical classification accuracy on different training setups.}
    \label{fig:hier_plot}
\end{figure}

{The results, shown in Fig.~\ref{fig:hier_plot}, compare the generalization performance of two training configurations when tested on outdoor data across three classification levels.}

{When the model is trained on indoor data only, it achieves moderate accuracy at the coarser levels (Modulation: 69.95\%, Protocol: 66.11\%) but struggles significantly with model-level identification (42.07\%). This indicates poor domain generalization from indoor to outdoor environments.}

{In contrast, training on a combined set of indoor and outdoor data yields substantial improvements across all hierarchical levels: accuracy increases to 89.42\% for Modulation, 87.34\% for Protocol, and 69.63\% for Model. The most significant gain (+27.56\,pp) is observed at the model level, the task most sensitive to environmental shifts. This demonstrates that training on diverse, mixed-domain data is crucial for learning robust, domain-invariant features.}

\subsection{Open-Set Recognition}
\subsubsection{Task Definition}
{Open-Set Recognition (OSR) extends classification to simultaneously (1)~accurately classify known drone types from the training set and (2)~identify and reject unknown samples as ``novel'' or ``outlier'' classes, preventing forced misclassification into known categories.}

\subsubsection{Baseline Model Architecture and Training}
{We utilize MetaMax~\cite{lyu2023metamax}, a state-of-the-art OSR method that combines deep feature learning with statistical modeling of known class distributions. For our OSR experiments, we curate a subset of CDRF comprising 14~distinct drone types totaling 2,180~spectrogram images. The dataset is strategically partitioned into known and unknown classes to simulate realistic deployment scenarios: the known class set contains 10~drone types with 1,915~images, and the unknown class set comprises 4~drone types with 265~images. The data are split into training (60\%) and validation (20\%) samples from known classes only, with the remaining samples reserved for testing.}

\subsubsection{Evaluation Metrics}
{Performance metrics include (1)~closed-set accuracy on known classes, (2)~unknown detection rate, and (3)~overall accuracy.}

\subsubsection{Experimental Results and Baseline Performance}
\begin{figure}[!ht]
    \centering
    \includegraphics[width=\columnwidth]{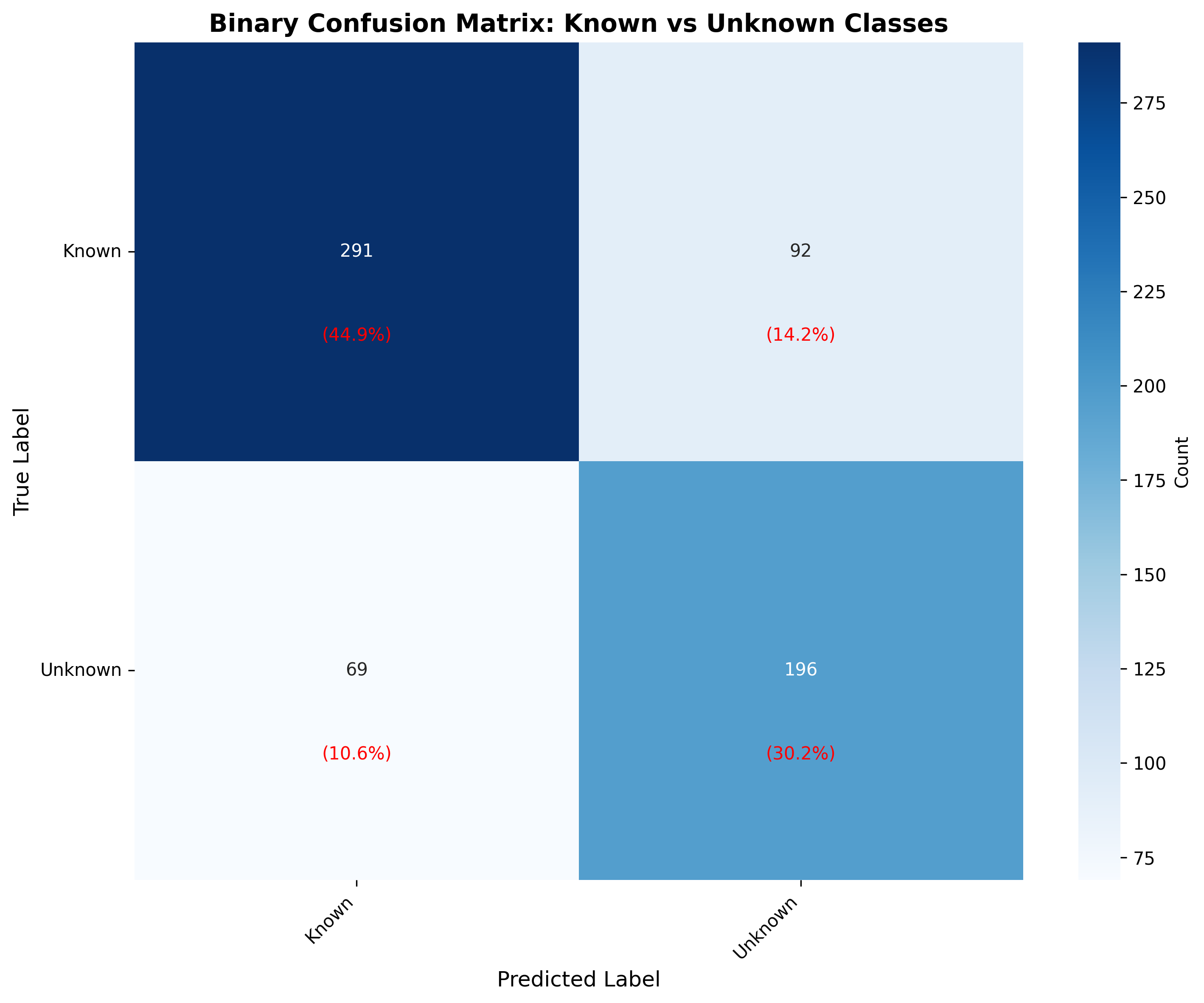}
    \caption{Binary confusion matrix for known vs unknown class discrimination.}
    \label{fig:osr_detailed_confusion}
\end{figure}

\begin{figure}[!ht]
    \centering
    \includegraphics[width=\columnwidth]{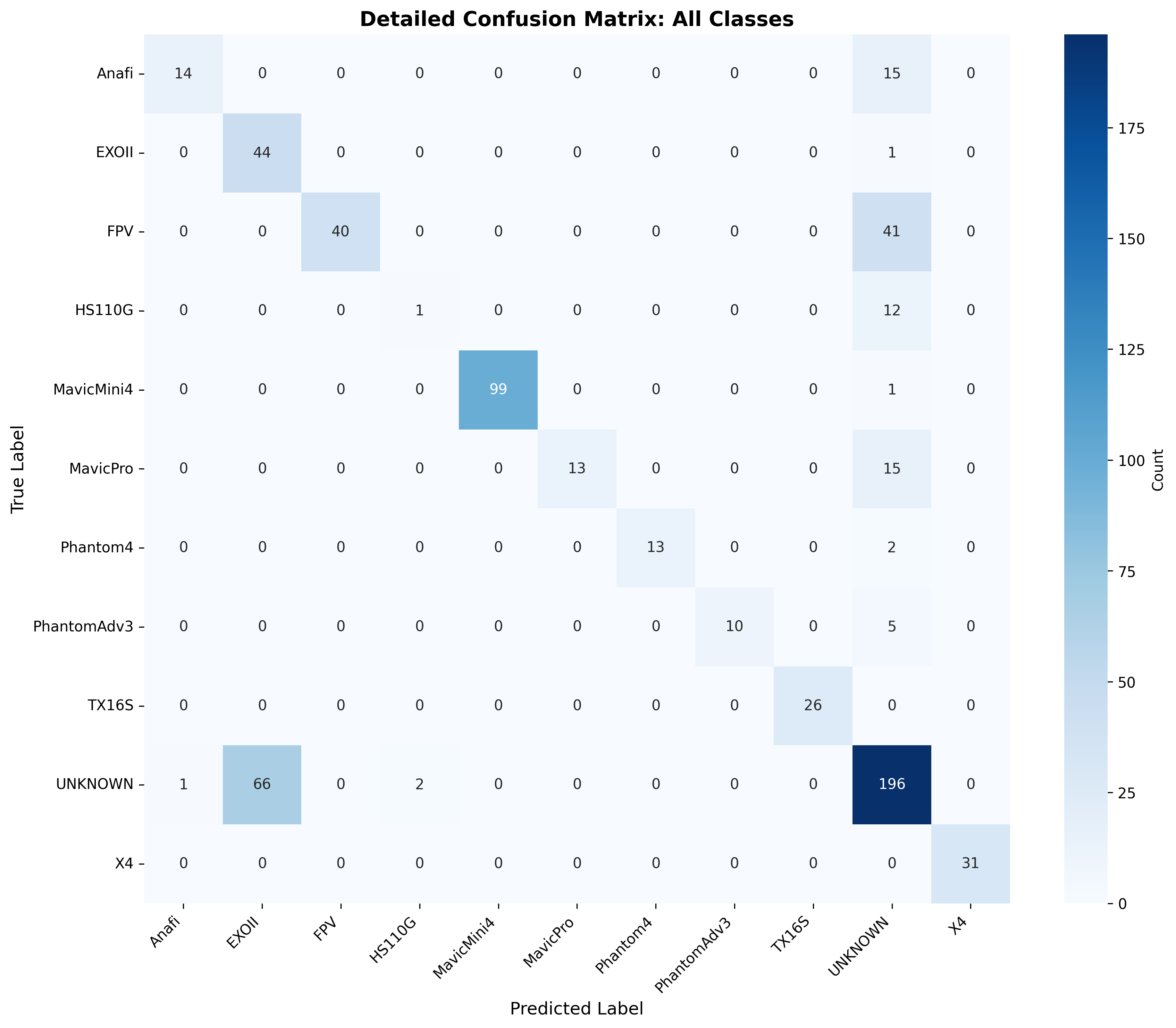}
    \caption{Detailed confusion matrix showing per-class performance for open-set recognition.}
    \label{fig:osr_binary_confusion}
\end{figure}

{MetaMax achieves an overall accuracy of 75.15\%, with known class accuracy of 75.98\% and an unknown detection rate of 73.96\%. The detailed confusion matrix (Fig.~\ref{fig:osr_detailed_confusion}) reveals a strong correlation between performance and class size. Among the 265~unknown test samples, 196 (73.96\%) are correctly identified as novel. The binary confusion matrix for known vs.\ unknown discrimination is shown in Fig.~\ref{fig:osr_binary_confusion}. These results demonstrate both the promise and challenges of OSR for RF drone detection.}

\subsection{Drone Classification (Single-Label)}
\subsubsection{Task Definition}
The single-label drone classification task involves assigning each detected RF signal to exactly one drone type from a predefined set of classes.

\subsubsection{Baseline Model Architecture and Training}
{We employ a ResNet-18 convolutional neural network as the backbone, leveraging its proven effectiveness for spectrogram-based RF signal classification. The model is trained on spectrogram images generated from the CDRF dataset using standard data augmentation and normalization techniques to improve generalization. The dataset contains 20~distinct drone types. The ResNet-18 model is trained with cross-entropy loss and optimized using Adam.}
 
\subsubsection{Experimental Results and Baseline Performance}

\begin{figure}[!ht]
    \centering
    \includegraphics[width=\columnwidth]{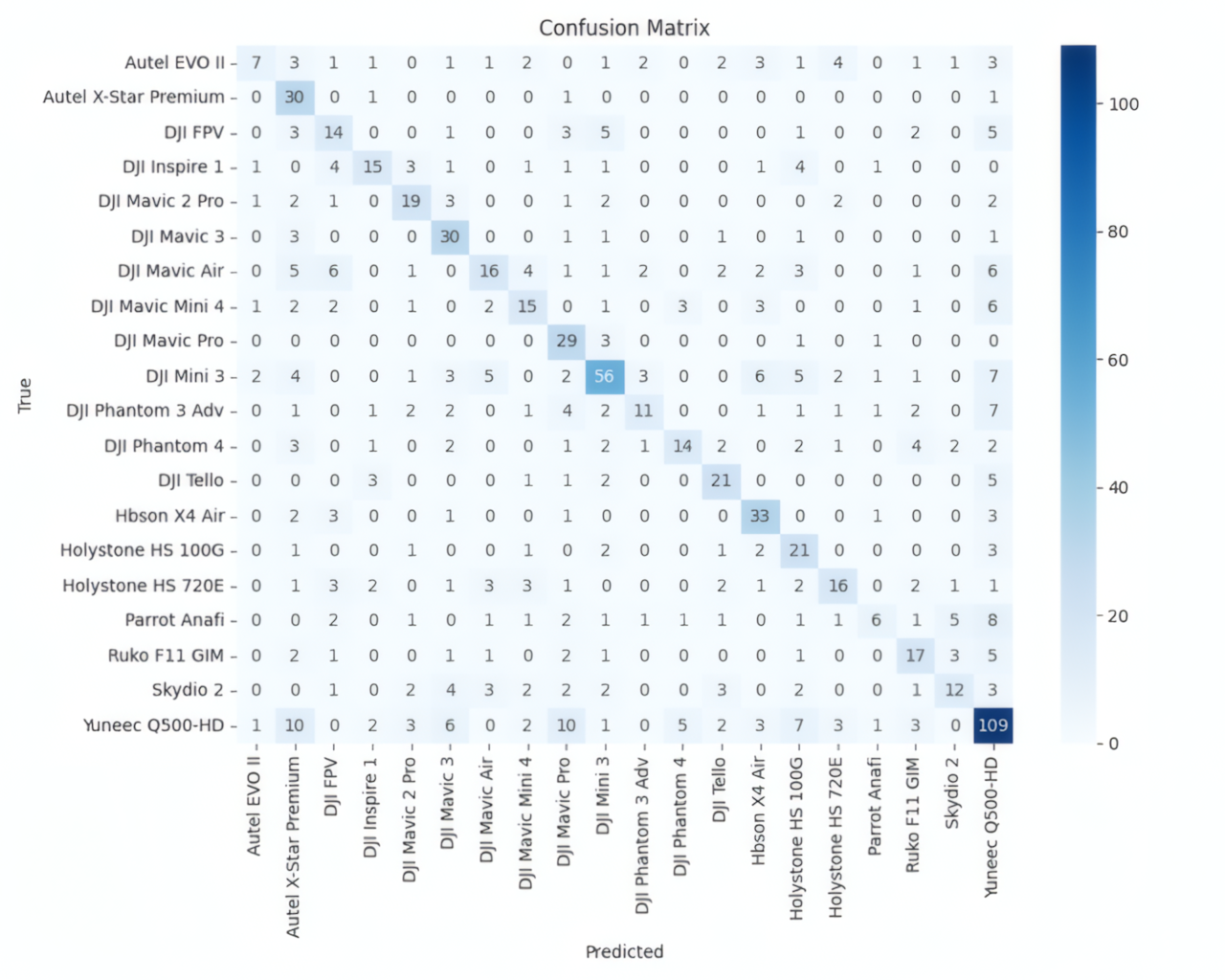}
    \caption{Confusion matrix for single-label drone classification.}
    \label{fig:flat_confusion_matrix}
\end{figure}

{The model achieves a test accuracy of 53.49\%, as shown in the confusion matrix (Fig.~\ref{fig:flat_confusion_matrix}). The confusion matrix reveals that certain drone types are classified with high accuracy, while others exhibit significant misclassifications, particularly for classes with fewer training samples. These results highlight the challenges of single-label classification of RF drone signals, especially when distinguishing between similar models.}

%% file: sections/5_conclusion.tex
\section{Discussion and Future Work}
\label{sec:discussion}
{The CDRF benchmark and its accompanying toolkit represent a significant step toward addressing the critical need for robust, reproducible, and realistic evaluation of RF-based drone perception systems. By providing a large-scale dataset rich in device diversity and environmental conditions, coupled with a principled raw-signal augmentation pipeline, CDRF enables the research community to move beyond idealized benchmarks and tackle the challenges of real-world deployments. The strong performance of baseline models under clean, controlled conditions underscores the data quality, while the performance degradation under augmentation highlights the gap between laboratory performance and operational robustness.}

{Our experiments reveal several key challenges and opportunities. The drop in recall for the YOLOv11n detector under augmented conditions suggests that, while the whole-signal annotation policy is effective, more advanced architectures or training strategies may be needed to maintain high sensitivity in low-SNR and high-interference scenarios. Similarly, although the open-set recognition baseline shows promise, reliably detecting unknown drone threats remains a significant challenge that requires further innovation in both model design and feature representation.}

{To ensure the long-term relevance and utility of CDRF, we plan to pursue several directions:}
\begin{itemize}
    \item {\textbf{Dataset expansion:} Regular updates with new and emerging drone models, more diverse background and interference signals from a wider range of environments (e.g., urban, industrial, and rural), and integration with other public datasets to create a more comprehensive benchmark.}
    \item {\textbf{Advanced model architectures:} Exploration of more sophisticated models, such as Transformers and graph neural networks, which may better capture the complex temporal and spectral dependencies in RF signals.}
    \item {\textbf{Enhanced augmentation techniques:} Development of more advanced augmentation techniques that more realistically simulate a wider range of real-world channel effects, such as multipath fading and Doppler shifts.}
    \item {\textbf{Advanced tasks:} Further investigation of multi-label and open-set classification, which are critical for real-world applications where multiple drones may be present and novel threats are a constant concern.}
\end{itemize}

\section{Conclusion}
\label{sec:conclusion}
{In this paper, we introduced CDRF, a comprehensive benchmark dataset and open-source toolkit for RF-based drone detection, classification, and open-set recognition. CDRF addresses key limitations of existing datasets by providing a large-scale, diverse collection of real-world and synthetically augmented RF signals, coupled with tools for reproducible data processing, augmentation, and evaluation. Our baseline experiments across a range of ML tasks demonstrate the utility of CDRF for developing and evaluating robust drone perception models. By open-sourcing the dataset and toolkit, we aim to foster a more collaborative and rigorous research environment, accelerating progress toward reliable, field-ready counter-drone systems.}

%% file: bibtex/bib/IEEEexample.bib
@misc{rfuav,
  title={RFUAV: A Benchmark Dataset for Unmanned Aerial Vehicle Detection and Identification},
  author={Shi, Rui and Yu, Xiaodong and Wang, Shengming and Zhang, Yijia and Xu, Lu and Pan, Peng and Ma, Chunlai},
  year={2025},
  eprint={2503.09033},
  archivePrefix={arXiv},
  primaryClass={cs.RO},
  url={https://arxiv.org/abs/2503.09033},
  note={arXiv preprint}
}

@article{robotics12020053,
AUTHOR = {Chen, Ziming and Yan, Jinjin and Ma, Bing and Shi, Kegong and Yu, Qiang and Yuan, Weijie},
TITLE = {A Survey on Open-Source Simulation Platforms for Multi-Copter UAV Swarms},
JOURNAL = {Robotics},
VOLUME = {12},
YEAR = {2023},
NUMBER = {2},
ARTICLE-NUMBER = {53},
URL = {https://www.mdpi.com/2218-6581/12/2/53},
ISSN = {2218-6581},
DOI = {10.3390/robotics12020053}
}

@article{seidaliyeva2023advances,
  title={Advances and challenges in drone detection and classification techniques: A state-of-the-art review},
  author={Seidaliyeva, Ulzhalgas and Ilipbayeva, Lyazzat and Taissariyeva, Kyrmyzy and Smailov, Nurzhigit and Matson, Eric T},
  journal={Sensors},
  volume={24},
  number={1},
  pages={125},
  year={2023},
  publisher={MDPI}
}

@inproceedings{teoh2019rf,
  title={RF and network signature-based machine learning on detection of wireless controlled drone},
  author={Teoh, Yan Jun John and Seow, Chee Kiat},
  booktitle={2019 PhotonIcs \& Electromagnetics Research Symposium-Spring (PIERS-Spring)},
  pages={408--417},
  year={2019},
  organization={IEEE}
}

@article{mrabet2024machine,
  title={Machine learning algorithms applied for drone detection and classification: benefits and challenges},
  author={Mrabet, Manel and Sliti, Maha and Ammar, Lassaad Ben},
  journal={Frontiers in Communications and Networks},
  volume={5},
  pages={1440727},
  year={2024},
  publisher={Frontiers Media SA}
}

@article{al2024deep,
  title={Deep learning for unmanned aerial vehicles detection: A review},
  author={Al-lQubaydhi, Nader and Alenezi, Abdulrahman and Alanazi, Turki and Senyor, Abdulrahman and Alanezi, Naif and Alotaibi, Bandar and Alotaibi, Munif and Razaque, Abdul and Hariri, Salim},
  journal={Computer Science Review},
  volume={51},
  pages={100614},
  year={2024},
  publisher={Elsevier}
}

@misc{bello2019radio,
  title={Radio frequency toolbox for drone detection and classification},
  author={Bello, Abdulkabir},
  year={2019},
  note={Unpublished/technical report}
}

@article{medaiyese2022wavelet,
  title={Wavelet transform analytics for RF-based UAV detection and identification system using machine learning},
  author={Medaiyese, Olusiji O and Ezuma, Martins and Lauf, Adrian P and Guvenc, Ismail},
  journal={Pervasive and Mobile Computing},
  volume={82},
  pages={101569},
  year={2022},
  publisher={Elsevier}
}

@article{nguyen2025effective,
  title={An Effective RF-Based Solution For Drone Detection And Recognition Amid Noise, Bluetooth, And Wi-Fi Interference},
  author={Nguyen, Trong Thanh and Nguyen, Le Cuong and Nguyen, Thi-Thanh-Tan},
  journal={Signal, Image and Video Processing},
  volume={19},
  number={9},
  pages={702},
  year={2025},
  publisher={Springer}
}

@article{kilicc2022drone,
  title={Drone classification using RF signal based spectral features},
  author={K{\i}l{\i}{\c{c}}, Rabiye and Kumbasar, Nida and Oral, Emin Argun and Ozbek, Ibrahim Yucel},
  journal={Engineering Science and Technology, an International Journal},
  volume={28},
  pages={101028},
  year={2022},
  publisher={Elsevier}
}

@article{basak2021combined,
  title={Combined RF-based drone detection and classification},
  author={Basak, Sanjoy and Rajendran, Sreeraj and Pollin, Sofie and Scheers, Bart},
  journal={IEEE Transactions on Cognitive Communications and Networking},
  volume={8},
  number={1},
  pages={111--120},
  year={2021},
  publisher={IEEE}
}

@inproceedings{allahham2020deep,
  title={Deep learning for RF-based drone detection and identification: A multi-channel 1-D convolutional neural networks approach},
  author={Allahham, Mhd Saria and Khattab, Tamer and Mohamed, Amr},
  booktitle={2020 IEEE International Conference on Informatics, IoT, and Enabling Technologies (ICIoT)},
  pages={112--117},
  year={2020},
  organization={IEEE}
}

@article{mo2022deep,
  title={Deep learning approach to UAV detection and classification by using compressively sensed RF signal},
  author={Mo, Yongguang and Huang, Jianjun and Qian, Gongbin},
  journal={Sensors},
  volume={22},
  number={8},
  pages={3072},
  year={2022},
  publisher={MDPI}
}

@article{fioranelli2015classification,
  title={Classification of loaded/unloaded micro-drones using multistatic radar},
  author={Fioranelli, Francesco and Ritchie, Matthew and Griffiths, H and Borrion, H},
  journal={Electronics Letters},
  volume={51},
  number={22},
  pages={1813--1815},
  year={2015},
  publisher={Wiley Online Library}
}

@article{mahjourian2024multimodal,
  title={Multimodal object detection using depth and image data for manufacturing parts},
  author={Mahjourian, Nazanin and Nguyen, Vinh},
  journal={arXiv preprint arXiv:2411.09062},
  year={2024}
}

@inproceedings{drozdowicz201635,
  title={35 GHz FMCW drone detection system},
  author={Drozdowicz, Jedrzej and Wielgo, Maciej and Samczynski, Piotr and Kulpa, Krzysztof and Krzonkalla, Jaroslaw and Mordzonek, Maj and Bryl, Marcin and Jakielaszek, Zbigniew},
  booktitle={2016 17th International Radar Symposium (IRS)},
  pages={1--4},
  year={2016},
  organization={IEEE}
}

@article{coluccia2020detection,
  title={Detection and classification of multirotor drones in radar sensor networks: A review},
  author={Coluccia, Angelo and Parisi, Gianluca and Fascista, Alessio},
  journal={Sensors},
  volume={20},
  number={15},
  pages={4172},
  year={2020},
  publisher={MDPI}
}

@article{zhu2020classification,
  title={Classification of UAV-to-ground targets based on enhanced micro-Doppler features extracted via PCA and compressed sensing},
  author={Zhu, Lingzhi and Zhang, Shuning and Ma, Qun and Zhao, Huichang and Chen, Si and Wei, Dongxu},
  journal={IEEE Sensors Journal},
  volume={20},
  number={23},
  pages={14360--14368},
  year={2020},
  publisher={IEEE}
}

@article{roldan2020dopplernet,
  title={DopplerNet: A convolutional neural network for recognising targets in real scenarios using a persistent range--Doppler radar},
  author={Roldan, Ignacio and del-Blanco, Carlos R and Duque de Quevedo, {\'A}lvaro and Iba{\~n}ez Urzaiz, Fernando and Gismero Menoyo, Javier and Asensio L{\'o}pez, Alberto and Berj{\'o}n, Daniel and Jaureguizar, Fernando and Garc{\'\i}a, Narciso},
  journal={IET Radar, Sonar \& Navigation},
  volume={14},
  number={4},
  pages={593--600},
  year={2020},
  publisher={Wiley Online Library}
}

@article{rahman2020classification,
  title={Classification of drones and birds using convolutional neural networks applied to radar micro-Doppler spectrogram images},
  author={Rahman, Samiur and Robertson, Duncan A},
  journal={IET radar, sonar \& navigation},
  volume={14},
  number={5},
  pages={653--661},
  year={2020},
  publisher={Wiley Online Library}
}

@inproceedings{moses2011radar,
  title={Radar-based detection and identification for miniature air vehicles},
  author={Moses, Allistair and Rutherford, Matthew J and Valavanis, Kimon P},
  booktitle={2011 IEEE international conference on control applications (CCA)},
  pages={933--940},
  year={2011},
  organization={IEEE}
}

@inproceedings{mendis2016deep,
  title={Deep learning based doppler radar for micro UAS detection and classification},
  author={Mendis, Gihan J and Randeny, Tharindu and Wei, Jin and Madanayake, Arjuna},
  booktitle={MILCOM 2016-2016 IEEE Military Communications Conference},
  pages={924--929},
  year={2016},
  organization={IEEE}
}

@article{andravsi2017night,
  title={Night-time detection of uavs using thermal infrared camera},
  author={Andra{\v{s}}i, Petar and Radi{\v{s}}i{\'c}, Tomislav and Mu{\v{s}}tra, Mario and Ivo{\v{s}}evi{\'c}, Jurica},
  journal={Transportation research procedia},
  volume={28},
  pages={183--190},
  year={2017},
  publisher={Elsevier}
}

@inproceedings{mezei2015drone,
  title={Drone sound detection},
  author={Mezei, J{\'o}zsef and Fiaska, Viktor and Moln{\'a}r, Andr{\'a}s},
  booktitle={2015 16th IEEE International Symposium on Computational Intelligence and Informatics (CINTI)},
  pages={333--338},
  year={2015},
  organization={IEEE}
}

@inproceedings{mezei2016drone,
  title={Drone sound detection by correlation},
  author={Mezei, J{\'o}zsef and Moln{\'a}r, Andr{\'a}s},
  booktitle={2016 IEEE 11th International Symposium on Applied Computational Intelligence and Informatics (SACI)},
  pages={509--518},
  year={2016},
  organization={IEEE}
}

@inproceedings{nijim2016drone,
  title={Drone classification and identification system by phenome analysis using data mining techniques},
  author={Nijim, Mais and Mantrawadi, Nikhil},
  booktitle={2016 IEEE Symposium on Technologies for Homeland Security (HST)},
  pages={1--5},
  year={2016},
  organization={IEEE}
}

@article{yue2018software,
  title={Software defined radio and wireless acoustic networking for amateur drone surveillance},
  author={Yue, Xuejun and Liu, Yongxin and Wang, Jian and Song, Houbing and Cao, Huiru},
  journal={IEEE Communications Magazine},
  volume={56},
  number={4},
  pages={90--97},
  year={2018},
  publisher={IEEE}
}

@article{bernardini2017drone,
  title={Drone detection by acoustic signature identification},
  author={Bernardini, Andrea and Mangiatordi, Federica and Pallotti, Emiliano and Capodiferro, Licia},
  journal={electronic imaging},
  volume={29},
  pages={60--64},
  year={2017},
  publisher={Society for Imaging Science and Technology}
}

@article{gokcce2015vision,
  title={Vision-based detection and distance estimation of micro unmanned aerial vehicles},
  author={G{\"o}k{\c{c}}e, Fatih and {\"U}{\c{c}}oluk, G{\"o}kt{\"u}rk and {\c{S}}ahin, Erol and Kalkan, Sinan},
  journal={Sensors},
  volume={15},
  number={9},
  pages={23805--23846},
  year={2015},
  publisher={MDPI}
}

@inproceedings{saqib2017study,
  title={A study on detecting drones using deep convolutional neural networks},
  author={Saqib, Muhammad and Khan, Sultan Daud and Sharma, Nabin and Blumenstein, Michael},
  booktitle={2017 14th IEEE international conference on advanced video and signal based surveillance (AVSS)},
  pages={1--5},
  year={2017},
  organization={IEEE}
}

@inproceedings{thomas2019uav,
  title={UAV localization using panoramic thermal cameras},
  author={Thomas, Anthony and Leboucher, Vincent and Cotinat, Antoine and Finet, Pascal and Gilbert, Mathilde},
  booktitle={International Conference on Computer Vision Systems},
  pages={754--767},
  year={2019},
  organization={Springer}
}

@inproceedings{park2015combination,
  title={Combination of radar and audio sensors for identification of rotor-type unmanned aerial vehicles (uavs)},
  author={Park, Seongha and Shin, Sangmi and Kim, Yongho and Matson, Eric T and Lee, Kyuhwan and Kolodzy, Paul J and Slater, Joseph C and Scherreik, Matthew and Sam, Monica and Gallagher, John C and others},
  booktitle={2015 IEEE SENSORS},
  pages={1--4},
  year={2015},
  organization={IEEE}
}

@inproceedings{liu2017drone,
  title={Drone detection based on an audio-assisted camera array},
  author={Liu, Hao and Wei, Zhiqiang and Chen, Yitong and Pan, Jie and Lin, Le and Ren, Yunfang},
  booktitle={2017 IEEE Third International Conference on Multimedia Big Data (BigMM)},
  pages={402--406},
  year={2017},
  organization={IEEE}
}

@inproceedings{kim2017real,
  title={Real-time UAV sound detection and analysis system},
  author={Kim, Juhyun and Park, Cheonbok and Ahn, Jinwoo and Ko, Youlim and Park, Junghyun and Gallagher, John C},
  booktitle={2017 IEEE Sensors Applications Symposium (SAS)},
  pages={1--5},
  year={2017},
  organization={IEEE}
}

@article{al2019rf,
  title={RF-based drone detection and identification using deep learning approaches: An initiative towards a large open source drone database},
  author={Al-Sa’d, Mohammad F and Al-Ali, Abdulla and Mohamed, Amr and Khattab, Tamer and Erbad, Aiman},
  journal={Future Generation Computer Systems},
  volume={100},
  pages={86--97},
  year={2019},
  publisher={Elsevier}
}

@inproceedings{al2020drone,
  title={Drone detection approach based on radio-frequency using convolutional neural network},
  author={Al-Emadi, Sara and Al-Senaid, Felwa},
  booktitle={2020 IEEE International Conference on Informatics, IoT, and Enabling Technologies (ICIoT)},
  pages={29--34},
  year={2020},
  organization={IEEE}
}

@inproceedings{akter2020rf,
  title={RF-based UAV surveillance system: A sequential convolution neural networks approach},
  author={Akter, Rubina and Doan, Van-Sang and Tunze, Godwin Brown and Lee, Jae-Min and Kim, Dong-Seong},
  booktitle={2020 International Conference on Information and Communication Technology Convergence (ICTC)},
  pages={555--558},
  year={2020},
  organization={IEEE}
}

@inproceedings{medaiyese2021machine,
  title={Machine learning framework for RF-based drone detection and identification system},
  author={Medaiyese, Olusiji O and Syed, Abbas and Lauf, Adrian P},
  booktitle={2021 2nd International Conference On Smart Cities, Automation \& Intelligent Computing Systems (ICON-SONICS)},
  pages={58--64},
  year={2021},
  organization={IEEE}
}

@article{taha2019machine,
  title={Machine learning-based drone detection and classification: State-of-the-art in research},
  author={Taha, Bilal and Shoufan, Abdulhadi},
  journal={IEEE access},
  volume={7},
  pages={138669--138682},
  year={2019},
  publisher={IEEE}
}

@article{nguyen2018cost,
  title={Cost-effective and passive rf-based drone presence detection and characterization},
  author={Nguyen, Phuc and Truong, Hoang and Ravindranathan, Mahesh and Nguyen, Anh and Han, Richard and Vu, Tam},
  journal={GetMobile: Mobile Computing and Communications},
  volume={21},
  number={4},
  pages={30--34},
  year={2018},
  publisher={ACM New York, NY, USA}
}

@article{fu2024radio,
  title={Radio Frequency Signal-Based Drone Classification with Frequency Domain Gramian Angular Field and Convolutional Neural Network.},
  author={Fu, Yuanhua and He, Zhiming},
  journal={Drones (2504-446X)},
  volume={8},
  number={9},
  year={2024}
}

@inproceedings{abeywickrama2018rf,
  title={RF-based direction finding of UAVs using DNN},
  author={Abeywickrama, Samith and Jayasinghe, Lahiru and Fu, Hua and Nissanka, Subashini and Yuen, Chau},
  booktitle={2018 IEEE International Conference on Communication Systems (ICCS)},
  pages={157--161},
  year={2018},
  organization={IEEE}
}

@inproceedings{shijith2017breach,
  title={Breach detection and mitigation of UAVs using deep neural network},
  author={Shijith, N and Poornachandran, Prabaharan and Sujadevi, VG and Dharmana, Meher Madhu},
  booktitle={2017 Recent Developments in Control, Automation \& Power Engineering (RDCAPE)},
  pages={360--365},
  year={2017},
  organization={IEEE}
}

@article{kim2016drone,
  title={Drone classification using convolutional neural networks with merged Doppler images},
  author={Kim, Byung Kwan and Kang, Hyun-Seong and Park, Seong-Ook},
  journal={IEEE Geoscience and Remote Sensing Letters},
  volume={14},
  number={1},
  pages={38--42},
  year={2016},
  publisher={IEEE}
}

@inproceedings{basak2021drone,
  title={Drone classification from RF fingerprints using deep residual nets},
  author={Basak, Sanjoy and Rajendran, Sreeraj and Pollin, Sofie and Scheers, Bart},
  booktitle={2021 International Conference on COMmunication Systems \& NETworkS (COMSNETS)},
  pages={548--555},
  year={2021},
  organization={IEEE}
}

@article{drone2024daiarticle,
author = {Dai, Xuanze},
year = {2024},
month = {03},
pages = {92-100},
title = {Drone detection with radio frequency signals and deep learning models},
volume = {47},
journal = {Applied and Computational Engineering},
doi = {10.54254/2755-2721/47/20241230}
}

@article{frid2024drones,
  title={Drones detection using a fusion of rf and acoustic features and deep neural networks},
  author={Frid, Alan and Ben-Shimol, Yehuda and Manor, Erez and Greenberg, Shlomo},
  journal={Sensors},
  volume={24},
  number={8},
  pages={2427},
  year={2024},
  publisher={MDPI}
}

@article{KILIC2022101028,
title = {Drone classification using RF signal based spectral features},
journal = {Engineering Science and Technology, an International Journal},
volume = {28},
pages = {101028},
year = {2022},
issn = {2215-0986},
doi = {https://doi.org/10.1016/j.jestch.2021.06.008},
url = {https://www.sciencedirect.com/science/article/pii/S2215098621001403},
author = {Rabiye Kılıç and Nida Kumbasar and Emin Argun Oral and Ibrahim Yucel Ozbek}
}

@inproceedings{podder2024deep,
  title={Deep learning for UAV detection and classification via Radio frequency signal analysis},
  author={Podder, Prajoy and Zawodniok, Maciej and Madria, Sanjay},
  booktitle={2024 25th IEEE International Conference on Mobile Data Management (MDM)},
  pages={165--174},
  year={2024},
  organization={IEEE}
}

@article{akter2021cnn,
  title={CNN-SSDI: Convolution neural network inspired surveillance system for UAVs detection and identification},
  author={Akter, Rubina and Doan, Van-Sang and Lee, Jae-Min and Kim, Dong-Seong},
  journal={Computer Networks},
  volume={201},
  pages={108519},
  year={2021},
  publisher={Elsevier}
}

@article{dadrass2022modified,
  title={A modified YOLOv4 Deep Learning Network for vision-based UAV recognition},
  author={Dadrass Javan, Farzaneh and Samadzadegan, Farhad and Gholamshahi, Mehrnaz and Ashatari Mahini, Farnaz},
  journal={Drones},
  volume={6},
  number={7},
  pages={160},
  year={2022},
  publisher={MDPI}
}

@inproceedings{kabir2021deep,
  title={Deep learning inspired vision based frameworks for drone detection},
  author={Kabir, Muhammad Salman and Ndukwe, Ikechi Kalu and Awan, Engr Zainab Shahid},
  booktitle={2021 International Conference on Electrical, Communication, and Computer Engineering (ICECCE)},
  pages={1--5},
  year={2021},
  organization={IEEE}
}

@article{casabianca2021acoustic,
  title={Acoustic-based UAV detection using late fusion of deep neural networks},
  author={Casabianca, Pietro and Zhang, Yu},
  journal={Drones},
  volume={5},
  number={3},
  pages={54},
  year={2021},
  publisher={MDPI}
}

@article{elyousseph2024robustness,
  title={Robustness of deep-learning-based RF UAV detectors},
  author={Elyousseph, Hilal and Altamimi, Majid},
  journal={Sensors},
  volume={24},
  number={22},
  pages={7339},
  year={2024},
  publisher={MDPI}
}

@article{gluge2024robust,
  title={Robust low-cost drone detection and classification using convolutional neural networks in low SNR environments},
  author={Gl{\"u}ge, Stefan and Nyfeler, Matthias and Aghaebrahimian, Ahmad and Ramagnano, Nicola and Sch{\"u}pbach, Christof},
  journal={IEEE Journal of Radio Frequency Identification},
  year={2024},
  publisher={IEEE}
}

@article{allahham2019dronerf,
  title={DroneRF dataset: A dataset of drones for RF-based detection, classification and identification},
  author={Allahham, MHD Saria and Al-Sa'd, Mohammad F and Al-Ali, Abdulla and Mohamed, Amr and Khattab, Tamer and Erbad, Aiman},
  journal={Data in brief},
  volume={26},
  pages={104313},
  year={2019}
}

@misc{5jjj-1m32-21,
  author = {Swinney, Carolyn J. and Woods, John C.},
  title = {DroneDetect Dataset: A Radio Frequency Dataset of Unmanned Aerial System (UAS) Signals for Machine Learning Detection \& Classification},
  year = {2021},
  publisher = {IEEE Dataport},
  doi = {10.21227/5jjj-1m32},
  url = {https://doi.org/10.21227/5jjj-1m32},
  note = {Dataset}
}

@misc{1xp7-ge95-22,
  author = {Medaiyese, Olusiji and Ezuma, Martins and Lauf, Adrian and Adeniran, Ayodeji},
  title = {Cardinal RF (CardRF): An Outdoor UAV/UAS/Drone RF Signals with Bluetooth and WiFi Signals Dataset},
  year = {2022},
  publisher = {IEEE Dataport},
  doi = {10.21227/1xp7-ge95},
  url = {https://doi.org/10.21227/1xp7-ge95},
  note = {Dataset}
}

@misc{VTI_DroneSET_FFT,
  author = {Sazdi\'c-Joti\'c, Boban and Pokrajac, Ivan and Bajcetic, Jovan and Bondzulic, Boban and Joksimovi\'c, Vasilija and \v{S}evi\'c, Tamara and Obradovic, Danilo},
  title = {{VTI\_DroneSET\_FFT}},
  year = {2021},
  month = jan,
  doi = {10.17632/s6tgnnp5n2.3},
  url = {https://doi.org/10.17632/s6tgnnp5n2.3},
  note = {Dataset}
}

@inproceedings{gluge2023robust,
  title={Robust drone detection and classification from radio frequency signals using convolutional neural networks},
  author={Gl{\"u}ge, Stefan and Nyfeler, Matthias and Ramagnano, Nicola and Horn, Claus and Sch{\"u}pbach, Christof},
  booktitle={15th International Joint Conference on Computational Intelligence (IJCCI), Rome, Italy, 13-15 November 2023},
  pages={496--504},
  year={2023},
  organization={SciTePress}
}

@inproceedings{akter2022explainable,
  title={An explainable multi-task learning approach for rf-based uav surveillance systems},
  author={Akter, Rubina and Doan, Van-Sang and Zainudin, Ahmad and Kim, Dong-Seong},
  booktitle={2022 Thirteenth International Conference on Ubiquitous and Future Networks (ICUFN)},
  pages={145--149},
  year={2022},
  organization={IEEE}
}

@article{jamil2020malicious,
  title={Malicious UAV detection using integrated audio and visual features for public safety applications},
  author={Jamil, Sonain and Fawad and Rahman, MuhibUr and Ullah, Amin and Badnava, Salman and Forsat, Masoud and Mirjavadi, Seyed Sajad},
  journal={Sensors},
  volume={20},
  number={14},
  pages={3923},
  year={2020},
  publisher={MDPI}
}

@inproceedings{svanstrom2021real,
  title={Real-time drone detection and tracking with visible, thermal and acoustic sensors},
  author={Svanstr{\"o}m, Fredrik and Englund, Cristofer and Alonso-Fernandez, Fernando},
  booktitle={2020 25th International Conference on Pattern Recognition (ICPR)},
  pages={7265--7272},
  year={2021},
  organization={IEEE}
}

@inproceedings{diamantidou2019multimodal,
  title={Multimodal deep learning framework for enhanced accuracy of UAV detection},
  author={Diamantidou, Eleni and Lalas, Antonios and Votis, Konstantinos and Tzovaras, Dimitrios},
  booktitle={International Conference on Computer Vision Systems},
  pages={768--777},
  year={2019},
  organization={Springer}
}

@inproceedings{jovanoska2018multisensor,
  title={Multisensor data fusion for UAV detection and tracking},
  author={Jovanoska, Snezhana and Br{\"o}tje, Martina and Koch, Wolfgang},
  booktitle={2018 19th international radar symposium (IRS)},
  pages={1--10},
  year={2018},
  organization={IEEE}
}

@inproceedings{lyu2023metamax,
  title={Metamax: improved open-set deep neural networks via weibull calibration},
  author={Lyu, Zongyao and Gutierrez, Nolan B and Beksi, William J},
  booktitle={Proceedings of the IEEE/CVF Winter Conference on Applications of Computer Vision},
  pages={439--443},
  year={2023}
}

@inproceedings{nelega2023radio,
  title={Radio frequency-based drone detection and classification using deep learning algorithms},
  author={Nelega, Raluca and Turcu, Romulus Valeriu Flaviu and Belean, Bogdan and Puschita, Emanuel},
  booktitle={2023 International Conference on Software, Telecommunications and Computer Networks (SoftCOM)},
  pages={1--6},
  year={2023},
  organization={IEEE}
}

@article{khanam2024yolov11,
  title={Yolov11: An overview of the key architectural enhancements},
  author={Khanam, Rahima and Hussain, Muhammad},
  journal={arXiv preprint arXiv:2410.17725},
  year={2024}
}

@inproceedings{redmon2016you,
  title={You only look once: Unified, real-time object detection},
  author={Redmon, Joseph and Divvala, Santosh and Girshick, Ross and Farhadi, Ali},
  booktitle={Proceedings of the IEEE conference on computer vision and pattern recognition},
  pages={779--788},
  year={2016}
}

@misc{roboflow_website,
  author       = {{Roboflow}},
  title        = {Roboflow: Computer Vision Development Platform},
  howpublished = {\url{https://roboflow.com}},
  note         = {Accessed: 2025-10-06}
}

@article{paszke2019pytorch,
  title={Pytorch: An imperative style, high-performance deep learning library},
  author={Paszke, Adam and Gross, Sam and Massa, Francisco and Lerer, Adam and Bradbury, James and Chanan, Gregory and Killeen, Trevor and Lin, Zeming and Gimelshein, Natalia and Antiga, Luca and others},
  journal={Advances in neural information processing systems},
  volume={32},
  year={2019}
}

@inproceedings{he2016deep,
  title={Deep residual learning for image recognition},
  author={He, Kaiming and Zhang, Xiangyu and Ren, Shaoqing and Sun, Jian},
  booktitle={Proceedings of the IEEE conference on computer vision and pattern recognition},
  pages={770--778},
  year={2016}
}

@article{grossmann1984decomposition,
  title={Decomposition of Hardy functions into square integrable wavelets of constant shape},
  author={Grossmann, Alexander and Morlet, Jean},
  journal={SIAM journal on mathematical analysis},
  volume={15},
  number={4},
  pages={723--736},
  year={1984},
  publisher={SIAM}
}

@book{cohen1995time,
  title     = {Time-Frequency Analysis: Theory and Applications},
  author    = {Cohen, Leon},
  year      = {1995},
  publisher = {Prentice Hall},
  address   = {Englewood Cliffs, NJ}
}

@article{bai2020intelligent,
  title={Intelligent diagnosis for railway wheel flat using frequency-domain Gramian angular field and transfer learning network},
  author={Bai, Yongliang and Yang, Jianwei and Wang, Jinhai and Li, Qiang},
  journal={Ieee Access},
  volume={8},
  pages={105118--105126},
  year={2020},
  publisher={IEEE}
}

@article{bruna2013invariant,
  title={Invariant scattering convolution networks},
  author={Bruna, Joan and Mallat, St{\'e}phane},
  journal={IEEE transactions on pattern analysis and machine intelligence},
  volume={35},
  number={8},
  pages={1872--1886},
  year={2013},
  publisher={IEEE}
}

@article{vaswani2017attention,
  title={Attention is all you need},
  author={Vaswani, Ashish and Shazeer, Noam and Parmar, Niki and Uszkoreit, Jakob and Jones, Llion and Gomez, Aidan N and Kaiser, {\L}ukasz and Polosukhin, Illia},
  journal={Advances in neural information processing systems},
  volume={30},
  year={2017}
}

@inproceedings{liu2021swin,
  title={Swin transformer: Hierarchical vision transformer using shifted windows},
  author={Liu, Ze and Lin, Yutong and Cao, Yue and Hu, Han and Wei, Yixuan and Zhang, Zheng and Lin, Stephen and Guo, Baining},
  booktitle={Proceedings of the IEEE/CVF international conference on computer vision},
  pages={10012--10022},
  year={2021}
}

@article{dosovitskiy2020image,
  title={An image is worth 16x16 words: Transformers for image recognition at scale},
  author={Dosovitskiy, Alexey and Beyer, Lucas and Kolesnikov, Alexander and Weissenborn, Dirk and Zhai, Xiaohua and Unterthiner, Thomas and Dehghani, Mostafa and Minderer, Matthias and Heigold, Georg and Gelly, Sylvain and others},
  journal={arXiv preprint arXiv:2010.11929},
  year={2020}
}

@inproceedings{tan2019efficientnet,
  title={Efficientnet: Rethinking model scaling for convolutional neural networks},
  author={Tan, Mingxing and Le, Quoc},
  booktitle={International conference on machine learning},
  pages={6105--6114},
  year={2019},
  organization={PMLR}
}

@article{howard2017mobilenets,
  title={Mobilenets: Efficient convolutional neural networks for mobile vision applications},
  author={Howard, Andrew G and Zhu, Menglong and Chen, Bo and Kalenichenko, Dmitry and Wang, Weijun and Weyand, Tobias and Andreetto, Marco and Adam, Hartwig},
  journal={arXiv preprint arXiv:1704.04861},
  year={2017}
}

@article{hochreiter1997long,
  title={Long short-term memory},
  author={Hochreiter, Sepp and Schmidhuber, J{\"u}rgen},
  journal={Neural computation},
  volume={9},
  number={8},
  pages={1735--1780},
  year={1997},
  publisher={MIT press}
}

@inproceedings{lea2016temporal,
  title={Temporal convolutional networks: A unified approach to action segmentation},
  author={Lea, Colin and Vidal, Rene and Reiter, Austin and Hager, Gregory D},
  booktitle={European conference on computer vision},
  pages={47--54},
  year={2016},
  organization={Springer}
}

@inproceedings{chen2016xgboost,
  title={Xgboost: A scalable tree boosting system},
  author={Chen, Tianqi and Guestrin, Carlos},
  booktitle={Proceedings of the 22nd acm sigkdd international conference on knowledge discovery and data mining},
  pages={785--794},
  year={2016}
}

@book{oppenheim1999discrete,
  title={Discrete-time signal processing},
  author={Oppenheim, Alan V},
  year={1999},
  publisher={Pearson Education India}
}

@article{virtanen2020scipy,
  title={SciPy 1.0: fundamental algorithms for scientific computing in Python},
  author={Virtanen, Pauli and Gommers, Ralf and Oliphant, Travis E and Haberland, Matt and Reddy, Tyler and Cournapeau, David and Burovski, Evgeni and Peterson, Pearu and Weckesser, Warren and Bright, Jonathan and others},
  journal={Nature methods},
  volume={17},
  number={3},
  pages={261--272},
  year={2020},
  publisher={Nature Publishing Group US New York}
}

@article{jocher2022ultralytics,
  title={ultralytics/yolov5: v7. 0-yolov5 sota realtime instance segmentation},
  author={Jocher, Glenn and Chaurasia, Ayush and Stoken, Alex and Borovec, Jirka and Kwon, Yonghye and Michael, Kalen and Fang, Jiacong and Yifu, Zeng and Wong, Colin and Montes, Diego and others},
  journal={Zenodo},
  year={2022}
}

@inproceedings{lin2014microsoft,
  title={Microsoft coco: Common objects in context},
  author={Lin, Tsung-Yi and Maire, Michael and Belongie, Serge and Hays, James and Perona, Pietro and Ramanan, Deva and Doll{\'a}r, Piotr and Zitnick, C Lawrence},
  booktitle={European conference on computer vision},
  pages={740--755},
  year={2014},
  organization={Springer}
}
